\documentclass{article}

\PassOptionsToPackage{numbers, compress}{natbib}
\usepackage{wrapfig,lipsum,booktabs}

\usepackage[preprint]{neurips_2024}
\usepackage{color,soul}

\usepackage[utf8]{inputenc} % allow utf-8 input
\usepackage[T1]{fontenc}    % use 8-bit T1 fonts
\usepackage{hyperref}       % hyperlinks
\usepackage{url}            % simple URL typesetting
\usepackage{booktabs}       % professional-quality tables
\usepackage{amsfonts}       % blackboard math symbols
\usepackage{nicefrac}       % compact symbols for 1/2, etc.
\usepackage{microtype}      % microtypography
\usepackage{xcolor}         % colors

\usepackage{graphicx}
\usepackage{multirow} % for borders and merged ranges
\usepackage{soul}% for underlines
\usepackage{amsmath}
\usepackage{algpseudocode}
\usepackage{makecell}
\usepackage{balance}
\usepackage{listings}
\usepackage{amsmath}
\usepackage{amssymb}
\usepackage{tcolorbox}
\usepackage{xspace}
\usepackage{pifont}
\newcommand{\dmel}{\texttt{dMel}\xspace}
\newcommand{\ourstts}{\texttt{RichTTS}\xspace}
\newcommand{\oursasr}{\texttt{RichASR}\xspace}

\usepackage{enumitem}
\soulregister{\xspace}{7}
\soulregister{\dmel}{7}
\soulregister{\cite}{7}
\newcommand{\xmark}{\ding{55}}%
\title{\dmel: Speech Tokenization made Simple}

\author{%
 Richard He Bai* \\
  Apple\\
  \texttt{richardbai@apple.com} \\
  \And
  Tatiana Likhomanenko* \\
  Apple \\
  \texttt{antares@apple.com} \\
  \And
  Ruixiang Zhang \\
  Apple \\
  \texttt{ruixiangz@apple.com} \\
  \And
  Zijin Gu \\
  Apple \\
    \texttt{zgu26@apple.com} \\
    \And
    Zakaria Aldeneh \\
    Apple \\
  \texttt{zaldeneh@apple.com} 
    \And
    Navdeep Jaitly \\
    Apple \\
  \texttt{njaitly@apple.com} 
}

\begin{document}
\def\thefootnote{*}\footnotetext{These authors contributed equally to this work.}
\def\thefootnote{\arabic{footnote}}

\maketitle

\begin{abstract}
  Large language models have revolutionized natural language processing by leveraging self-supervised pretraining on vast textual data. Inspired by this success, researchers have investigated various compression-based speech tokenization methods to discretize continuous speech signals, enabling the application of language modeling techniques to discrete tokens. However, audio compressor introduces additional complexity and computational cost, and often fail on out-of-domain audio signals.
In this work, we introduce a novel speech representation (\dmel) that discretizes mel-filterbank channels into intensity bins, creating a simpler yet more effective representation compared to existing speech tokenization methods. 
\textbf{Our approach demonstrates superior performance in preserving audio content, robustness to out-of-domain data, and offers a training-free, natural, and streamable representation.}
To address the high-dimensional nature of log-mel spectrograms, we propose an efficient parallel encoding and decoding method for high-dimensional tokens using an LM-style transformer architecture. This innovation enables us to develop \textbf{RichTTS} and \textbf{RichASR}—two models sharing the same architecture while achieving comparable or better results than specialized existing methods.
Our results demonstrate the effectiveness of \dmel in achieving high performance on both speech synthesis and recognition tasks within a unified framework, paving the way for efficient and effective joint modeling of speech and text. The code is available at \href{https://github.com/apple/dmel}{https://github.com/apple/dmel} and demos are available at \href{https://apple.github.io/dmel-demo/}{https://apple.github.io/dmel-demo/}.
\end{abstract}
\section{Introduction}
Large language models (LLMs) have achieved remarkable success in various natural language processing tasks by leveraging self-supervised pretraining on massive amounts of textual data~\citep{brown2020language}. 
Inspired by this success, numerous works~\citep{borsos2023audiolm, rubenstein2023audiopalm,zhang2023speechgpt,wang2023neural} have sought to extend the language modeling approach to speech processing, aiming to build unified models capable of both speech understanding and generation tasks.
However, a key challenge lies in the continuous nature of speech signals, necessitating effective tokenization methods to discretize the input for language model-based processing.

Current speech tokenization approaches can be broadly categorized into two types: semantic (content) tokens and acoustic tokens\footnote{We use a word `semantic' with the meaning of `content' to keep prior work notation \citep{borsos2023audiolm}.}. 
Semantic tokens, extracted from self-supervised (SSL) pretrained speech models~\citep{baevski2020wav2vec, hsu2021hubert}, are obtained by first encoding the speech signal into representations and then clustering them into discrete tokens with $k$-means method.
However, such SSL pretrained models are not useful for high fidelity speech synthesis as speaker identity and other details of raw speech are lost in training~\citep{borsos2023audiolm}.
Conversely, acoustic tokens can be obtained from audio compression models that are trained to compress the speech signal into codebook indices with residual vector quantization~(RVQ) and reconstruction objectives~\citep{zeghidour2021soundstream,defossez2022high}.
These tokens prioritize acoustic reconstruction but lose semantic information which can lead to poorer results in generating audio~\citep{wang2023neural}.

To combine the advantages of both semantic and acoustic tokens, AudioLM~\citep{borsos2023audiolm} proposed to model both semantic tokens and acoustic tokens with 3 stages: semantic modeling, coarse acoustic modeling, and fine acoustic modeling. 
The coarse-to-fine modeling strategy is designed to match the residual structure of RVQ based acoustic tokens.
This solution addresses both content and speech quality, but its multi-stage hierarchical structure complicates the model and can lead to slower training and inference.
Another solution is to combine the semantic and acoustic features together. \citet{zhang2023speechtokenizer} proposed to distill the semantic tokens into the acoustic token's first residual channel during the training of the RVQ model in a teacher-student manner.
In this way, the new feature can preserve the semantic information better and also reconstruct high quality speech signals.

\begin{figure}[!t]
  \centering
  \includegraphics[width=0.8\textwidth]{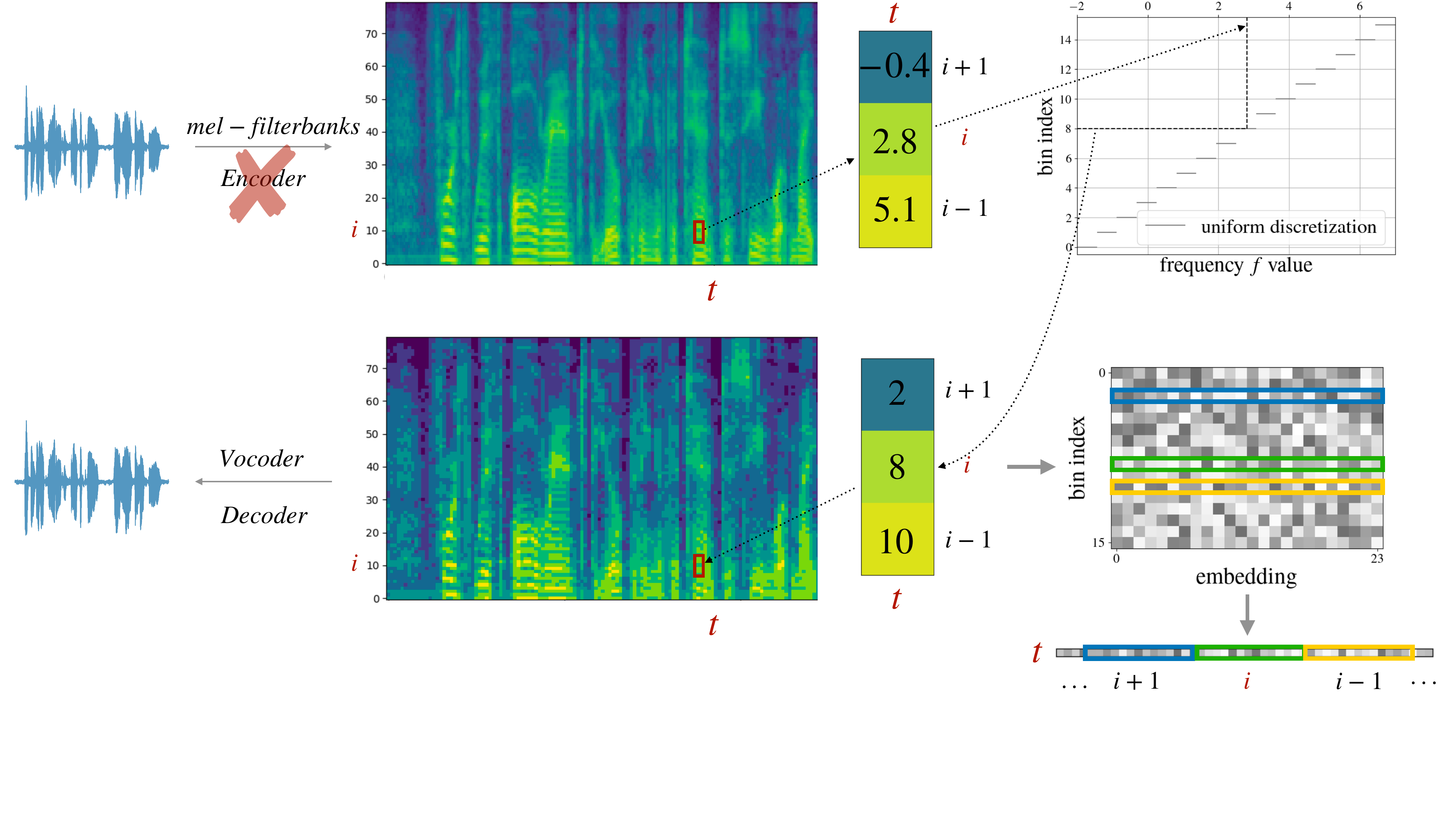}
  \vspace{-1.0cm}
  \caption{Prior works on speech tokenization use either heavy self-supervised pretrained encoders~\citep{baevski2020wav2vec, hsu2021hubert} to extract semantic tokens (and train a separate decoder for it~\citep{lakhotia2021generative}) or learn compression encoder-decoder models with residual vector quantizations~\citep{zeghidour2021soundstream,defossez2022high} to obtain acoustic tokens. By contrast we eliminate the encoder and simply discretize mel-filterbanks (\dmel) to encode audio, and use a simple mel-filterbank vocoder~\citep{yamamoto2020parallel} to reconstruct speech signals.}
  \label{fig:dmel}
\end{figure}

In this paper, we raise the following fundamental question -- \textbf{\textit{do we really need to separate speech into semantic and acoustic tokens first, and process them with idiosyncratic architectures?}}
We propose a simple alternative called \dmel (see Figure~\ref{fig:dmel}) that discretizes log mel-filterbanks (Mel) energies directly into ordinal bins. Intriguingly, we find that discretizing Mel has little impact on the ability of off-the-shelf Mel vocoders to reconstruct waveforms\footnote{We used vocoders from \url{https://github.com/kan-bayashi/ParallelWaveGAN}.}.
In Table~\ref{tab:teaser} we show different vocoders to reconstruct waveforms from Mel and discretized Mel (\dmel) computed on them, as well as ASR models trained on Mel and \dmel.

\begin{wraptable}{r}{7.5cm}
  \vspace{-0.4cm}
  \scriptsize
  \caption{Impact of discretization.}\label{tab:teaser}
  \begin{tabular}{lcc|cc}\toprule
    & \multicolumn{2}{c|}{Reconstruction WER (\%)} & \multicolumn{2}{c}{Recognition WER (\%)} \\
    & \textbf{P-WaveGAN\textsuperscript{1}} & \textbf{HifiGAN\textsuperscript{2}} & \textbf{Seq2seq\textsuperscript{3}} & \textbf{CTC\textsuperscript{4}} \\
    \midrule
    Ground-truth & \multicolumn{2}{c|}{2.02} & \multicolumn{2}{c}{-} \\
    \midrule
    Mel & 2.13 & 2.08 & 2.4 & 2.1 \\
    \dmel & 2.23 & 2.11 & 2.5 & 2.1 \\
  \bottomrule
  \end{tabular}
  \scriptsize
  \textsuperscript{*} 
\cite{yamamoto2020parallel}\textsuperscript{1}, \cite{kong2020hifi}\textsuperscript{2}, \cite{dong2018speech}\textsuperscript{3}, \cite{graves2006connectionist}\textsuperscript{4} Configurations are detailed in Sec.~\ref{sec:exp}.
\vspace{-0.2cm}
\end{wraptable}

We find that the word error rate (WER) of an ASR system run on the reconstructed waveforms is quite similar to the WER of the same system run on the ground-truth audio, showing that \dmel captures the acoustic information needed to reconstruct good waveforms. Similarly, we find that the WER of ASR models trained on Mel and \dmel are similar, indicating that \dmel effectively preserves semantic content. This demonstrates that discretizing Mel has limited impact on information content while providing the benefits of discrete tokens.

By operating on the log mel-filterbanks and preserving the frequency and intensity information (with some loss of resolution from discretization), \dmel \textit{inherently preserves both semantic and acoustic information in a unified representation}, without the need for separate tokenization or additional pretraining of a tokenization model. The key advantages of \dmel include: 
\vspace{-0.2cm}
\begin{itemize}[leftmargin=0.3cm]
    \item \textbf{Interpretable and complete}: Preserves both semantic and acoustic information in an interpretable physics-based representation, with minimal information loss from discretization.
    \item \textbf{Model-free and versatile}: Directly compatible with any mel-filterbank vocoder for waveform reconstruction, unlike other methods where representations are tightly coupled to specific encoder-decoder architectures.
    \item \textbf{Parallel processing}: Frequency channels can be \textbf{\textit{modeled independently}} without complex hierarchical dependencies, enabling efficient processing with decoder-only transformer architectures.
\end{itemize}
\vspace{-0.2cm}

While data-driven tokenization methods can be improved with more diverse training data, they inherently suffer from information loss due to their neural compression—discarding information that may be crucial in unseen conditions. In contrast, mel-spectrogram offers advantages as a physics-based signal representation:
\vspace{-0.2cm}
\begin{itemize}[leftmargin=0.3cm]
  \item \textbf{Robust frequency preservation}: Captures frequency components through principled transformation, aligning with human auditory perception's sensitivity to magnitude spectrum.
  \item \textbf{Domain-agnostic performance}: Demonstrates consistent behavior across acoustic conditions without requiring domain-specific training.
  \item \textbf{Proven reliability}: Shows robust performance across decades of speech processing research in diverse conditions.
\end{itemize}
\vspace{-0.2cm}

Given the high-dimensional nature of log-mel spectrograms, we propose an efficient parallel encoding and decoding method for these high-dimensional tokens using an LM-style transformer architecture. This enables us to develop RichTTS and RichASR—two models sharing the same architecture while achieving comparable or better results than specialized existing methods.

Through comprehensive evaluations, we show that using \dmel allows us to employ a single decoder-only model to achieve high performance on both automatic speech recognition ASR and TTS tasks.
The ASR task validates that \dmel preserves semantic information, while the TTS task demonstrates that \dmel is effective for high-fidelity acoustic reconstruction of speech.
We also compare \dmel{} to other tokenization methods and find that \dmel{} achieves the best WER for the ASR task, which indicates that semantic information is well preserved.
Additionally, \dmel{} achieves a lower WER score for the TTS task when using WhisperX~\citep{bain2022whisperx} for automatic evaluation, and we find that models trained with \dmel{} can generate long and natural speech samples.

\section{Method}

In this section, we first introduce our proposed \dmel speech tokenization method, which discretizes log mel-filterbanks energies directly into bins. We then describe our unified LM-style transformer model for ASR and TTS tasks, which leverages \dmel for speech tokenization. The model architecture is illustrated in Figure~\ref{fig:arch}.

\subsection{\dmel Speech Tokenizer}
Different from existing VQ-VAE~\citep{borsos2023audiolm,zhang2023speechtokenizer,kim2024clamtts, zeghidour2021soundstream} based speech tokenizers, we propose a discretized log mel-filterbanks based speech tokenizer.
The outline of the discretization method is shown in Figure~\ref{fig:dmel}.
Later in the paper, we show that this tokenizer allows the model to process the input speech signal efficiently and capture the relevant acoustic features for both ASR and TTS tasks.

We denote tensors as $\mathbf{X}$ while $\mathbf{X}_{i, ...}$ denote the $(i, ...)$-th component of tensor  $\mathbf{X}$. First, the speech tokenizer takes the input speech signal $\mathbf{x}$ and computes the log mel-filterbanks representation $\mathbf{M}$:
\begin{equation}
    \mathbf{M} = \text{Mel}(\mathbf{x}),
\end{equation}
where $\text{Mel}(\cdot)$ represents the function that computes the log mel-filterbanks,  $\mathbf{M}\in\mathbb{R}^{T\times N}$, $N$ is the number of log mel-filterbanks and $T$ is the number of frames in the spectrogram.
\vspace{-0.2cm}
\paragraph{Tokenization}
To discretize the log mel-filterbanks representation $\mathbf{M}$ into speech tokens, we adopt a codebook $\mathbf{C}$. 
In this paper, we apply a simple linear discretization, so that the codebook $\mathbf{C}\in\mathbb{R}^{2^K}$ and its values are evenly spaced in the range of the log mel-filterbanks values:
\begin{equation}
   m=\min_{t, i}(\mathbf{M}_{t, i}), 
   \qquad
   M=\max_{t, i}(\mathbf{M}_{t, i})
   \qquad
   \delta=\frac{M - m}{2^K},
\end{equation}
\begin{equation}
   \mathbf{C} = \left[ m,\, m + \delta, \, m + 2 \delta,\,  \dots, \, m + (2^K - 1) \delta \right].
\end{equation}
In practice, we compute the minimum $m$ and maximum $M$ values of log mel-filterbanks across the entire dataset to define the codebook $\mathbf{C}$.
Then we map a magnitude $\mathbf{M}_{t, i}$ of every frequency channel $i=1\dots N$ for the time frame $t=1\dots T$ into a bin index of the codebook $\mathbf{C}$ in the following way: 
% \begin{eqnarray}
%   \mathbf{S}_{t,i} &=& \text{Discretize}(\mathbf{M}_{t, i}) \\
%                 &=& \text{argmin}_j |\mathbf{M}_{t, i} - \mathbf{C}_j|
% \end{eqnarray}
\begin{equation}
\mathbf{S}_{t,i} = \text{Discretize}(\mathbf{M}_{t, i}) 
                = \text{argmin}_j |\mathbf{M}_{t, i} - \mathbf{C}_j|    
\end{equation}
where $\mathbf{S}\in\mathbf{B}^{T\times N}$ represents the discretized log mel-filterbanks (\dmel) with $\mathbf{B}=\{j|j=1,2,3, \dots 2^K\}$ and $\mathbf{S}_{t}\in\mathbf{B}^{N}$ being the $t$-th speech token. 
As the codebook $\mathbf{C}$ has $2^K$ distinct values and thus number of bins $|\mathbf{B}|=2^K$, each speech token is represented by $N\cdot K$ bits where every $K$ bits are used to represent one of $N$ frequency channels.
\vspace{-0.2cm}
\paragraph{Detokenization}
To reconstruct the speech signal $\mathbf{x}$ from the speech tokens $\mathbf{S}$, we first transform bin indices back to the log mel-filterbanks representation via the codebook $\mathbf{C}$:
\begin{equation}
    \hat{\mathbf{M}}_{t,i} = \mathbf{C}_{\mathbf{S}_{t,i}}.
\end{equation}
Then, we apply a vocoder~\citep{yamamoto2020parallel} to transform reconstructed log mel-filterbanks $\hat{\mathbf{M}}_{t,i}$ back into the time domain signal $\mathbf{x}$. 
The vocoder is trained independently and is not part of the transformer decoder-based model.

\paragraph{Comparison between Speech Tokenizers}
\begin{table}[!tp]\centering
  % \vspace{-0.3cm}
  \caption{Comparison between different speech tokenizers: \dmel (ours), HuBERT-KM and SpeechTokenizer. 
  For \dmel we use $N=80$ log mel-filterbanks (50ms window, 25ms hop distance), and $2^K=16$ values of the codebook $\mathbf{C}$. 
  For HuBERT-KM, 200 is chosen accoring to~\cite{maiti2024voxtlm}.
  }\label{tab:tokenizer}
  % \vspace{0.2cm}
  \small
  \begin{tabular}{lccc}\toprule
  &\dmel &HuBERT-KM &SpeechTokenizer \\\midrule
  Codebook Size &16 &200 &1024 \\
  Code Dimension &80 &1 &8 \\
  Vocabulary Size & 16 * 1 & 200 * 1 & 1024 * 8 \\
  Frame-rate & 40Hz & 50Hz & 50Hz \\
  Bit-rate & 12.8kps & 0.4kps & 4kps \\
  Training-free? &\checkmark & \ding{55} & \ding{55} \\
  \bottomrule
  \end{tabular}
  % \vspace{-0.4cm}
\end{table}

In Table\ref{tab:tokenizer}, we compare \dmel with the baselines in terms of vocabulary size, bit-rate, and frame-rate.
First, \dmel has a much smaller vocabulary, as it is discretized mel-filterbanks energies, allowing all 80 channels to share the same vocabulary since they represent similar energy values. 
In contrast, neural compression encoders like SpeechTokenizer require separate embeddings for different channels.
Also, \dmel operates at a lower frame-rate while maintaining a higher bit-rate.
The reduced frame-rate leads to shorter sequence lengths during both training and inference, which is particularly advantageous when using large models.
While a higher bit-rate typically increases model complexity for compression-based tokenizers, this is not the case for \dmel due to two key factors: i) \dmel is encoder-free, without any compression encoder;  ii) the complexity of the model introduced in Section~\ref{sec:model} depends only on the vocabulary size and sequence length (frame-rate), not on the code dimensions.
In compression-based methods, increasing the bit-rate requires either larger codebooks or additional residual dimensions, leading to increased tokenizer complexity. Moreover, these methods require more complex downstream models to handle the expanded representations.
Given recent studies \citep{defossez2024moshi,mousavi2024dasb} demonstrated that bit-rate does not strongly correlate with downstream model performance, we focus our comparative analysis on frame-rate rather than bit-rate when evaluating different speech tokens for downstream tasks. This approach challenges the conventional assumption that higher bit-rates necessarily yield better results.

\subsection{Unified Speech-Text Transformer Decoder (RichTTS and RichASR)}
\label{sec:model}

\begin{figure}[!t]
  \begin{minipage}[c]{0.5\textwidth}
      \hspace{-4.4cm} \includegraphics[width=1.4\textwidth]{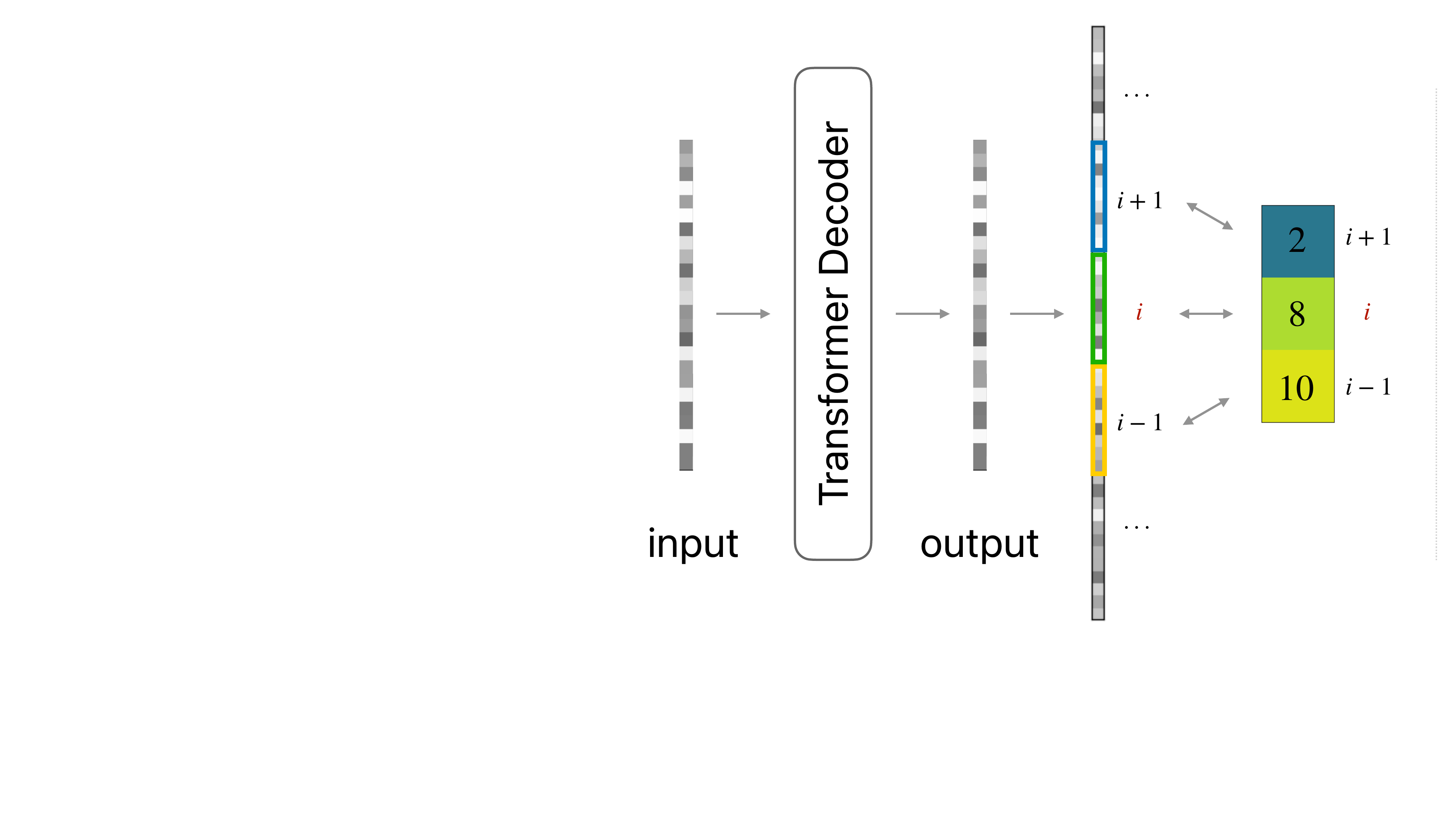}
  \end{minipage}
  \begin{minipage}[c]{0.5\textwidth}
    \hspace{-1.6cm} \includegraphics[width=1.2\textwidth]{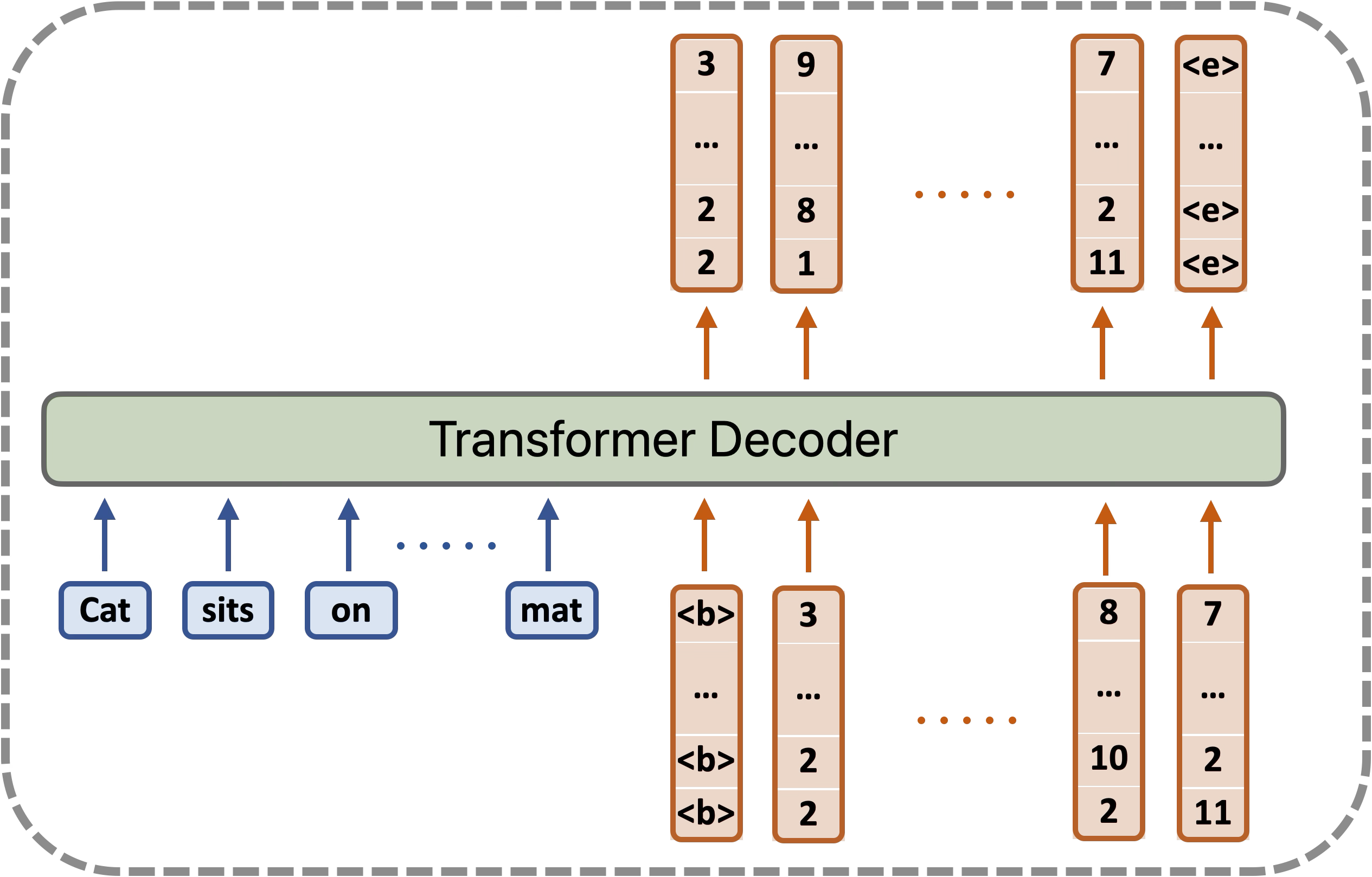}
  \end{minipage}
  % \vspace{-1.3cm}
  \caption{(Left) For a time step $t$ encoded \dmel from Figure~\ref{fig:dmel} is inputted to the transformer decoder to produce final embeddings for each of the frequency channels \textbf{in parallel}. (Right) Unified Speech-Text Transformer Decoder with speech tokens as \dmel{}.}
  \label{fig:arch}
\end{figure}

Modeling speech and text sequences jointly is essential for a model to understand and generate both modalities. 
However, it is challenging to design a unified model that can handle both speech-to-text and text-to-speech effectively.
In this work, we apply a unified LM-style transformer model that takes speech and text tokens as input and generates the output tokens in the target sequence.
The model is trained in end-to-end on a combined dataset of speech and text pairs, enabling it to learn the joint representations for ASR and TTS tasks. 
As we show in the rest of the paper, the crucial part for the joint model training is the proper speech tokenization which \dmel provides.
\vspace{-0.2cm}
\paragraph{Text Encoding}
For text data, we apply a character-level tokenizer to convert the input text into a sequence of text tokens. 
The text tokens are passed through an embedding layer, $\text{Embed}(\cdot): \{j|j=1,2,3\dots L\} \to \mathbb{R}^{D}$, where $D$ is the embedding dimension and $L$ is the vocabulary size.
The dimension of the speech token embedding is set to be the same as the text token embedding $D$ and no further mapping is required.
The motivation for using a character-level tokenizer is to reduce the vocabulary size $L$ and improve the model's generalization ability.
Also, character tokens can capture the fine-grained linguistic features that are essential for both ASR and TTS tasks.

% \paragraph{Speaker Encoding}
% To properly model multi-speaker data, we also include speaker embeddings as input to the transformer decoder. 
% The speaker embeddings are extracted from an independent dvector~\citep{variani2014deep} model\footnote{We use a pretrained model ``Speaker Encoder'' from the YourTTS~\citep{casanova2022yourtts} repository \url{https://github.com/Edresson/YourTTS}.}.
% We use a learnable linear layer to map the speaker embeddings to the same dimension as the speech and text token embeddings $D$.
% The speaker representation is optional for ASR task, but required for TTS task.
% Hence, during the training, it is applied for text-to-speech and ignored for speech-to-text.

\paragraph{Speech Encoding}
For speech signal, we apply the \dmel speech tokenizer to convert the input speech signal into a sequence of speech tokens. Then, the speech tokens $\mathbf{S}\in\mathbf{B}^{T\times N}$ are  passed through a learnable embedding layer, $\text{Embed}(\cdot): \mathbf{B} \to \mathbb{R}^{d}$, and a learnable linear layer, $\text{Linear}(\cdot): \mathbb{R}^{N \times d} \to \mathbb{R}^{D}$, to obtain the speech token representation $\mathbf{E}\in\mathbb{R}^{T\times D}$:
\begin{eqnarray}
  \mathbf{E}_{t}=\text{Linear}(\mathbf{E^{\prime}}_{t}), \, \text{and}\,\,\,
\mathbf{E^{\prime}}_{t}=\text{Concatenate}([\text{Embed}(\mathbf{S}_{t,1}), \text{Embed}(\mathbf{S}_{t,2}), \ldots, \text{Embed}(\mathbf{S}_{t,N})]),
\end{eqnarray}
where $\mathbf{E}_t\in\mathbb{R}^D$ is the speech token representation.
Here, for every time frame $t$, a speech token $\mathbf{S}_{t}$ is processed \textit{in parallel and independently} for every frequency channel $i$ by $\text{Embed}(\mathbf{S}_{t,i})$ mapping, and then  embeddings of all frequency channels are stacked together to form one vector representation $\mathbf{E^{\prime}}_{t}$ for the frame~$t$.
Finally, the speech token embeddings $\mathbf{E}_t$ are fed into the LM-style transformer models for further processing.

We also implemented other popular speech tokenizers including HuBERT-KM~\citep{lakhotia2021generative} and SpeechTokenizer~\citep{zhang2023speechtokenizer} for comparison.
The main difference among these speech tokenizers is the codebook size and codes dimension, shown in Table~\ref{tab:tokenizer}. 
For both HuBERT-KM and SpeechTokenizer the speech tokens are mapped via a learnable linear layer from their dimension to the text embedding dimension $D$ before feeding into the LM-style transformer model.
% \vspace{-0.4cm}

\begin{figure}[!t]
  \centering
  \includegraphics[width=0.5\textwidth]{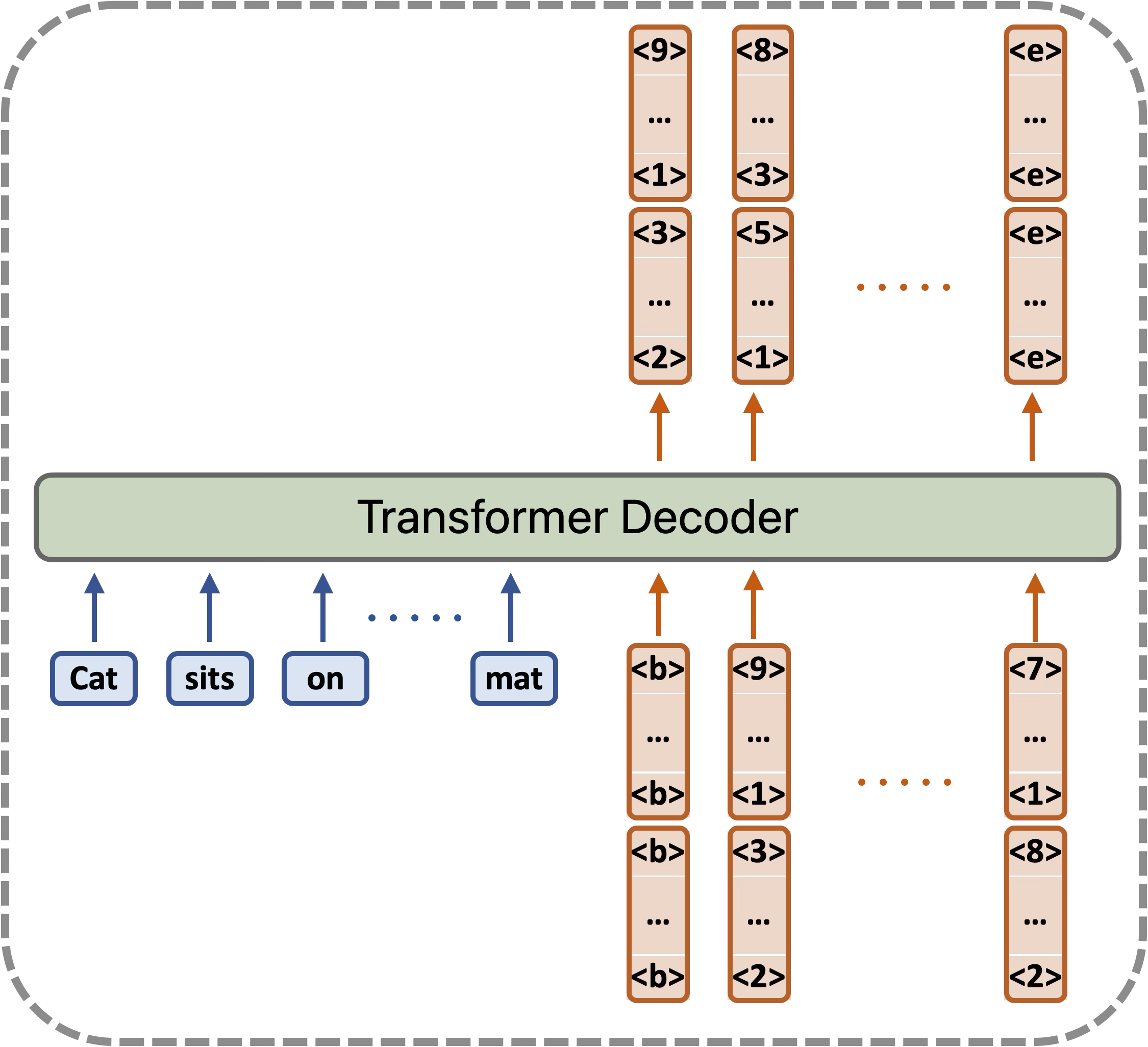}
  % \vspace{-1.8cm}
  \caption{Unified Speech-Text Transformer Decoder with speech tokens as \dmel{} where we predict multiple, e.g. two, frames in parallel reducing the frame rate, e.g. by 2x: \dmel tokens for every two frames are stacked together to form the input into the decoder and predicted in parallel afterwards.}
  \label{fig:arch-multiple-frames}
\end{figure}

\vspace{-0.2cm}
\paragraph{Transformer Decoder}
The transformer decoder is trained end-to-end on a combined dataset of speech and text pairs.
For TTS training, the input sequence is constructed by concatenating the speaker embedding (The speaker embeddings are extracted from an independent dvector~\citep{variani2014deep} model\footnote{We use a pretrained model ``Speaker Encoder'' from the YourTTS~\citep{casanova2022yourtts} repository \url{https://github.com/Edresson/YourTTS}.}), text tokens, and speech tokens. 
For ASR training, the input sequence is constructed by concatenating the speech tokens and text tokens.
Both tasks are trained with causal masking, where the model is trained to predict the next token based on the previous tokens.
The loss is calculated using the cross-entropy loss between the predicted tokens and the ground-truth tokens. Loss calculation is skipped on the speech tokens for ASR task and on the text tokens for TTS task. \textbf{\textit{Note, that all frequency channels at time frame $t$ for \dmel tokenizer are predicted independently and in parallel, see Figure~\ref{fig:arch} (left).}}
For positions encoding, to capture the relative distances between tokens in the input sequence, we apply multiplicative relative positional embedding RoPE~\citep{su2024roformer}. 
This allows the model to learn the positional relationships between speech tokens, text tokens, and speaker embeddings, enhancing its ability to generate coherent output sequences.
For positional embeddings we do not distinguish between text, speech and speaker tokens and thus having global positions notation across all of them, see Figure~\ref{fig:arch}.
% \vspace{-0.4cm}
\vspace{-0.2cm}
\paragraph{Context Masking during Training}
Compared to LMs, audio frames are highly redundant with strong local correlations.
This makes longform generation difficult for models due to exposure bias~\citep{scheduled_sampling}. 
To mitigate exposure bias during training, we apply span-masking~\citep{raffel2020exploring} to the speech token context, masking out multiple random spans of speech frames. 
The model is trained to predict the next token based on the masked context. 
This context-masking strategy helps the model learn to generate accurate speech tokens in the presence of missing information, improving its robustness and generalization. It forces the model to attend to the text rather than copying previously inferred speech tokens due to learnt correlations. We also find that span-masking text tokens improves the ASR task.

\paragraph{k-Frame Encoding and Decoding}
Given the fact that our model can encode and decode multiuple channels in parallel, for mel-spectrogram, we are insterested in whether the model can do k-frame encoding and decoding in parallel too.
As shown in Figure~\ref{fig:arch-multiple-frames}, we simply concatenate multiple frames together and using the same technique to encode and decode it as described in previous paragraphs.
Our results show that this model is also working well on multi-frame generation in parall. 
More results can be found in Section~\ref{sec:exp}.
% \vspace{-0.2cm}
% \paragraph{Predicting Multiple Frames in Parallel} 
% {\color{red} Richard TBD, check Figure~\ref{fig:arch-multiple-frames}.}

\section{Experiments}
\label{sec:exp}
In this section, we begin by evaluating different speech tokenizers through a common practice in the literature: tokenizing speech into discrete units and then reconstructing the speech to assess the quality of the reconstruction. 
This approach helps gauge the effectiveness of various tokenization techniques.
Following this, we present both TTS and ASR results using an LM-style (decoder-only) model with different speech tokens. 
While most related work focuses solely on speech synthesis, our study encompasses both speech generation and recognition, providing a more comprehensive evaluation of the tokenization methods.
We evaluate the performance of our model mainly on the LibriSpeech dataset and compare it with state-of-the-art speech tokenizers, ASR and TTS models.

\subsection{Speech Reconstruction on Clean Speech}
We first conduct speech reconstruction experiments with various speech tokenizers on clean speech. 
Following \cite{zhang2023speechtokenizer}, we randomly sample 300 speech utterances and their ground truth transcriptions from the LibriSpeech \textit{test-clean} dataset.
We use the speech2unit and unit2speech modules to convert the speech signal to speech tokens and then reconstruct the speech signal from the speech tokens.
We compute the WER between the ASR outputs from HuBERT-Large~\citep{hsu2021hubert}\footnote{We use checkpoint \url{https://huggingface.co/facebook/hubert-large-ls960-ft}.} on the audio samples and their ground truth transcripts. 
We also report MOS-LQO~(Mean Opinion Score -- Listening Quality Objective) score to measure the reconstruction quality using ViSQOL~\citep{hines2012visqol}.
Finally, we use human evaluation to measure the naturalness of the reconstructed speech using a MOS score with 95\% confidence interval.
We instruct the human evaluators to rate the naturalness of the reconstructed speech on a scale of 1 to 5, where 1 is the worst and 5 is the best.
The results are shown in Table~\ref{tab:reconstruction}.

% \begin{table}[!t]\centering
%   \caption{Speech reconstruction results on 300 random samples from LibriSpeech \textit{test-clean} set. 
%   For each tokenizer, we list its frame rate and the model parameter sizes of its speech2unit and unit2speech modules. WER is evaluated with HuBERT-Large and MOS-LQO with ViSQOL.}\label{tab:reconstruction}
%   \small
%   \begin{tabular}{@{}lrrcccccc}\toprule
%     \textbf{Tokenizer}&\textbf{Speech2unit} &\textbf{Unit2speech} &\textbf{Frame Rate} &\textbf{WER} &\textbf{MOS-LQO} &\textbf{MOS~(95\%Confidence)} \\\midrule
%     GroundTruth & -& -& -&2.02 &- &3.91$\pm$0.12 \\\midrule
%     HuBERT-KM~\cite{lakhotia2021generative} &95M &111M &50Hz &8.71 &2.06 &2.74$\pm$0.14 \\
%     EnCodec~\cite{defossez2022high} &7M &7M &75Hz &2.03 &4.03 &3.69$\pm$0.13 \\
%     SpeechTokenizer~\cite{zhang2023speechtokenizer} &65M &34M &50Hz &2.41 &4.19 &3.77$\pm$0.13 \\\midrule
%     \midrule
%     Mel-HifiGAN~\cite{kong2020hifi} &n/a &12M &80Hz &2.08 &4.52 &3.80$\pm$0.12 \\
%     \dmel-HifiGAN &n/a &12M &80Hz &2.11 &4.47 &3.68$\pm$0.13 \\
%     \midrule
%     Mel-PWG~\cite{yamamoto2020parallel} &n/a &1M &80Hz &2.13 &4.40 & 3.27$\pm$0.14\\
%     \dmel-PWG &n/a &1M &80Hz &2.23 &4.37 & 3.23$\pm$0.14\\
%     \midrule
%     Mel-PWG &n/a &1M &40Hz &2.36 &4.34 &2.99$\pm$0.15 \\
%     \dmel-PWG &n/a &1M &40Hz &2.51 &4.29 &2.97$\pm$0.15 \\
%     \bottomrule
%   \end{tabular}
%   \vspace{-0.4cm}
% \end{table}
\begin{table}[!t]
\centering
\vspace{-0.2cm}
\caption{Speech reconstruction results on 300 random samples from LibriSpeech \textit{test-clean} set. WER~(\%) is evaluated with Hubert Large.}\label{tab:reconstruction}
\vspace{0.2cm}
\resizebox{0.9\textwidth}{!}{
\begin{tabular}{@{}lrrcccc@{}}
\toprule
\textbf{Tokenizer} & \textbf{\begin{tabular}[c]{@{}c@{}}Speech2Unit\\ (M params)\end{tabular}} & \textbf{\begin{tabular}[c]{@{}c@{}}Unit2Speech\\ (M params)\end{tabular}} & \textbf{\begin{tabular}[c]{@{}c@{}}Frame\\ Rate\end{tabular}} & \textbf{WER↓} & \textbf{MOS-LQO↑} & \textbf{\begin{tabular}[c]{@{}c@{}}MOS↑\\ (95\% CI)\end{tabular}} \\
\midrule
GroundTruth & - & - & - & 2.02 & - & 3.91$\pm$0.12 \\
\midrule
HuBERT-KM & 95 & 111 & 50Hz & 8.71 & 2.06 & 2.74$\pm$0.14 \\
EnCodec & 7 & 7 & 75Hz & 2.03 & 4.03 & 3.69$\pm$0.13 \\
SpeechTokenizer & 65 & 34 & 50Hz & 2.41 & 4.19 & 3.77$\pm$0.13 \\
% Mimi & 37 & 38 & 12.5Hz & 2.33 & & \\
\midrule
Mel-HifiGAN & n/a & 12 & 80Hz & 2.08 & 4.52 & 3.80$\pm$0.12 \\
\dmel-HifiGAN & n/a & 12 & 80Hz & 2.11 & 4.47 & 3.68$\pm$0.13 \\
\midrule
Mel-PWG & n/a & 1 & 80Hz & 2.13 & 4.40 & 3.27$\pm$0.14 \\
\dmel-PWG & n/a & 1 & 80Hz & 2.23 & 4.37 & 3.23$\pm$0.14 \\
\midrule
Mel-PWG & n/a & 1 & 40Hz & 2.36 & 4.34 & 2.99$\pm$0.15 \\
\dmel-PWG & n/a & 1 & 40Hz & 2.51 & 4.29 & 2.97$\pm$0.15 \\
\bottomrule
\end{tabular}
}
% \vspace{-0.4cm}
\end{table}

\begin{figure}[!t]
  \centering
  \includegraphics[width=0.8\textwidth]{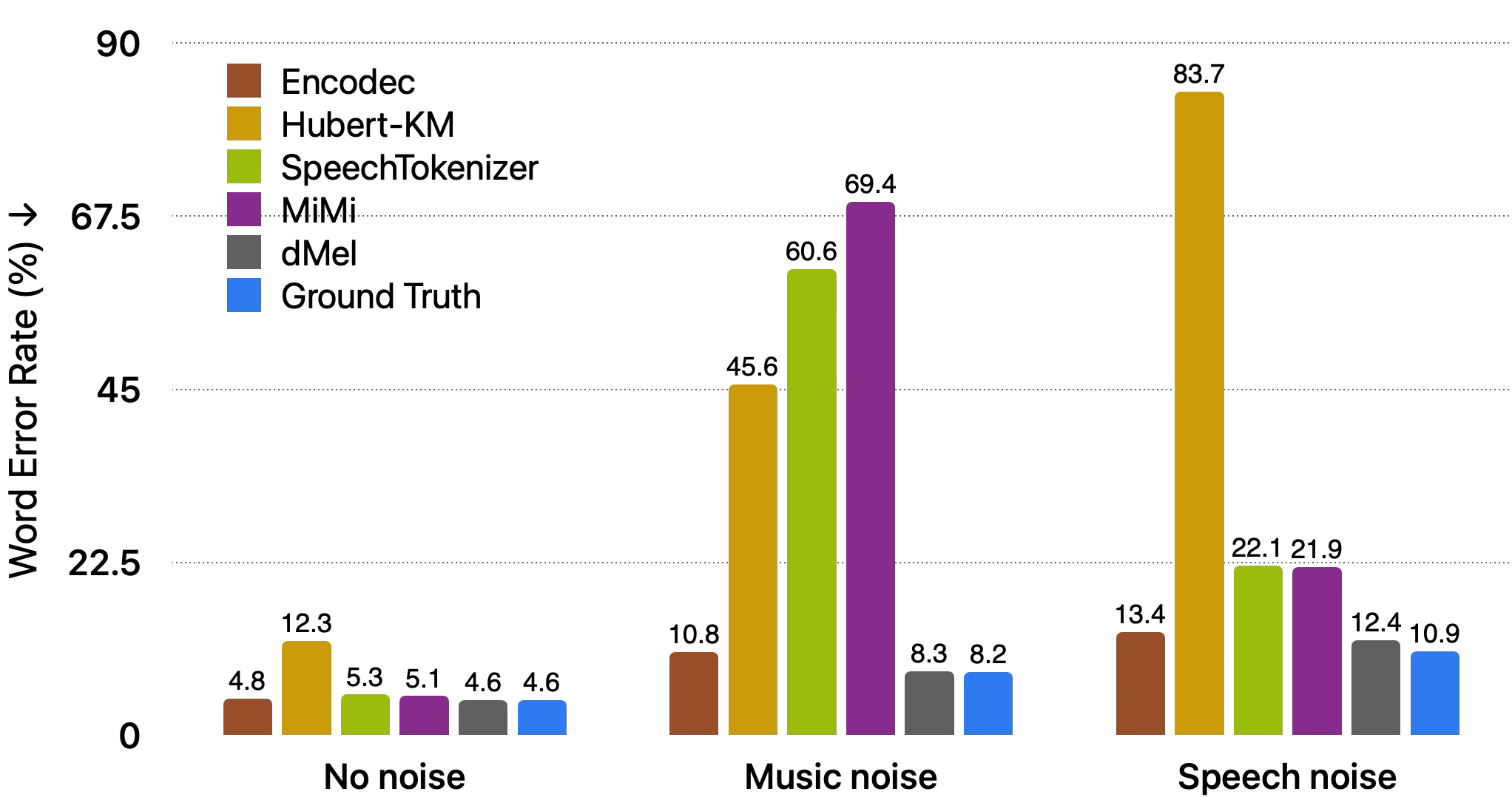}
  \caption{Speech reconstruction results on 300 random samples from LibriSpeech \textit{test-clean} set when noise is added: either background music from~\citep{bogdanov2019mtg} dataset or speech noise from \textit{test-other}. WER~(\%) is evaluated with WhisperX ASR (``base.en''). Audio examples are in our \href{https://apple.github.io/dmel-demo/}{demo page}.}
  \label{fig:reconstruction-noisy}
\end{figure}

From Table~\ref{tab:reconstruction}, we can see that semantic tokenization (HuBERT-KM) is poor for speech reconstruction. 
%as it applies a TTS model~(Tacotron2) and a vocoder~(WaveGlow) to reconstruct the speech signal. 
Meanwhile, acoustic tokenizers that are optimized to reconstruct the signal directly (EnCodec and SpeechTokenizer) do well.
%We find SpeechTokenizer is better than EnCodec in terms of human evaluation on naturalness, but its WER is higher.

We apply different vocoders to reconstruct the speech signal from log mel-filterbanks, and find that the WER of the reconstructed speech signal is comparable to the acoustic tokenization methods with a fraction of the parameters. 
Also, log mel-filterbanks achieve a better MOS-LQO score, which indicates that the reconstructed audio is more similar to the original audio.
By comparing Mel and \dmel, we can see that discretization has little impact on WER and MOS-LQO scores.
We also find that the exact vocoder matters much less than the frame rate of tokenization: the WER goes from 2.08 to 2.13 when switching from HifiGAN to ParallelWaveGAN (PWG), but it falls from 2.13 to 2.36 when the frame rate is changed from 80Hz to 40Hz.
However, even a 1M parameter vocoder operating at a 40Hz frame rate is comparable to the much larger SpeechTokenizer on WER and MOS-LQO metrics.

\subsection{Speech Reconstruction on Noisy Speech}
In addition to the in-domain clean human speech, we also measure speech tokenizer's out-of-domain ability which is neglected in prior work.
We compare different speech tokenizations for speech reconstruction under noisy conditions.
We syntetically add two kinds of noise signal to the clean audio: i) add music background to the original human speech; ii) add another human speech in low volume as a background to the original human speech.
Results are shown in Figure~\ref{fig:reconstruction-noisy}: both HuBERT-KM and SpeechTokenizer fail in out-of-domain setting while EnCodec, Mel and \dmel show robustness for noisy speech reconstruction. This supports our motivation to explore \dmel, training-free and determenistic tokenization, which is able to handle various acoustic conditions.

\textit{Considering the efficiency and performance, we choose the \dmel speech tokenizer in 40Hz with ParallelWaveGAN vocoder for the following experiments.}

\subsection{LM-Style Text-to-Speech}
  Here we compare the accuracy and naturalness of speech synthesized by LM-style text-to-speech~(TTS) models trained on different tokenization methods.
  For TTS evaluation, we utilize WhisperX~\citep{bain2022whisperx} (``base.en'' from~\cite{radford2023robust}) to transcribe our generated speech into text and calculate the WER and the character error rate~(CER).
  We report both WER and CER to facilitate comparisons to prior works which have reported only one or the other.
  \subsubsection{Configurations} \label{sec:tts_config}
    We use several open-sourced datasets with paired speech and text transcription to conduct experiments: i) LibriSpeech~\citep{panayotov2015librispeech} dataset (CC BY 4.0) consists of English speech recordings (960h, 16kHz) from various speakers ($\sim$2k) and conditions; ii) LibriTTS~\citep{zen2019libritts} (CC BY 4.0) dataset (500h) derived from LibriSpeech improves on it with the proper sentence split, text normalization and keeping samples 24kHz; iii) VCTK~\citep{vctk2019} contains 44h of English speech (108 speakers); iv) LJSpeech~\citep{ljspeech17} (public domain in US) is a single speaker English audio recordings of 22kHz with read speech from LibriVox\footnote{\url{https://librivox.org/pages/public-domain/}.}.
    While LibriSpeech is used to train ASR and TTS models, LibriTTS, VCTK and LJSpeech are only used to train the TTS.

    We train the LM-style transformers in three different sizes: Small, Base, and Large (see Appendix Table~\ref{tab:model-size}).
    Unless stated otherwise, the Base model is used in all experiments if not stated otherwise.
    All models use pre-LayerNorm with dropout set to 0.1 for residual, attention and embedding layers and 0.3 for positional embedding.
    \dmel uses 16 discrete bins for each channel while text is tokenized with a character vocabulary;  the speaker embedding dvector has 512 dimensions (see Appendx~\ref{app:training-details} for details). In all experiments, training data are sampled to 16kHz.

      We trained the TTS model using the same architecture but with three different tokenization methods: HuBERT+KM (with 200 clusters), SpeechTokenizer, and \dmel.
  Additionally, we present the results from VOXTLM~\citep{maiti2024voxtlm} and USLM~\citep{zhang2023speechtokenizer} for comparison. 
  VOXTLM is a larger model trained on more data that is initialized from a pretrained LLM~(OPT) using HuBERT-KM as the speech tokenizer. 
  USLM comprises an autoregressive (AR) model and a non-autoregressive (NAR) model, both trained with the SpeechTokenizer.
\subsubsection{Results}

  As shown in Table~\ref{tab:main-tts} for training on LibriSpeech dataset, our LM-style model with \dmel tokenization achieves a WER of 4.3 and a CER of 1.8, significantly outperforming the baseline methods. This indicates that our model can generate more accurate speech with less hallucination and distortion.
  Furthermore, we observed that the AR model trained on SpeechTokenizer tokens  exhibits a much higher WER compared to the idiosyncratic coarse to fine models (labeled AR+NAR) developed for these residual tokenizers -- indicating that \dmel lies on a simpler data manifold.

  Given the success of our LM-style \dmel TTS model, dubbed \ourstts, we further evaluate it on various datasets, including LJSpeech, VCTK, and LibriTTS, and compare it with popular open-sourced TTS models, including Tacotron2~\citep{shen2018natural}, FastSpeech2~\citep{ren2020fastspeech}, and VITS~\citep{kim2021conditional}.
  We conduct human evaluation to measure the naturalness of 50 randomly sampled synthesized speech from VCTK test set.
  \ourstts achieves competitive performance on the TTS task in terms of both MOS and WER demonstrating its effectiveness in generating high-quality synthesized speech, see Table~\ref{tab:tts}.
  Interestingly, we find that VITS performs poorly on the VCTK WER. 
  We suspect this is because VITS tends to make more mistakes at the beginning of each sequence, and since VCTK comprises short sequences, even one or two word errors can lead to a high WER.

%   \begin{table}[!t]\centering
%   \caption{Text-to-speech results for different tokenizers. All models are trained on LibriSpeech 960h dataset. Evaluation is done via speech generation on the full \textit{test-clean} transcriptions and speakers, and then evaluated WER with WhisperX base.en. We use \dag~ to annotate results from the reference paper. \ourstts is our model trained with \dmel tokenization. }\label{tab:main-tts}
%   \small
%   \begin{tabular}{lrrr}\toprule
%   &WER (\%) & CER (\%) \\\midrule
%   \ourstts (HuBERT+KM), 258M &9.5 &4.3 \\
%   VOXTLM (HuBERT+KM)\dag~\cite{maiti2024voxtlm}, 350M &- &3.5 \\
%   \ourstts (SpeechTokenizer), AR, 258M &11.4 &5.9 \\
%   USLM (SpeechTokenizer)\dag, AR+NAR~\cite{zhang2023speechtokenizer}, 356M &6.5 &- \\
%   \ourstts (\dmel), 258M &\textbf{4.3} &\textbf{1.8} \\
%   \bottomrule
% \end{tabular}
% % \vspace{-0.4cm}
% \end{table}

\begin{table}[!t]
\centering
\caption{Text-to-speech results for different tokenizers. \ourstts is trained on LibriSpeech 960h. WER (\%) and CER (\%) are evaluated with WhisperX ASR (``base.en'') and reported on \textit{test-clean}.}\label{tab:main-tts}
\vspace{0.2cm}
\resizebox{0.8\textwidth}{!}{
\begin{tabular}{@{}lrrc@{}}
\toprule
\textbf{Model} & \textbf{WER↓} (\%) & \textbf{CER↓} (\%) & \textbf{Params} \\
\midrule
VOXTLM (HuBERT+KM)\dag,~\cite{maiti2024voxtlm} & - & 3.5 & 350M \\
USLM (SpeechTokenizer)\dag, AR+NAR,~\cite{zhang2023speechtokenizer} & 6.5 & - & 356M \\
\midrule
\ourstts (HuBERT+KM) & 9.5 & 4.3 & 258M \\
\ourstts (SpeechTokenizer), AR & 11.4 & 5.9 & 258M \\
\ourstts (\dmel) & \textbf{4.3} & \textbf{1.8} & 258M \\
\bottomrule
\end{tabular}
}
% \vspace{-0.4cm}
\end{table}

% \begin{table}[!t]\centering
%   \caption{WER (\%) (evaluated with WhisperX base.en ASR model) of different TTS models' generations using transcriptions from the evaluation set of each dataset. Each column corresponds to the training and evaluation dataset. We used ESP-Net~\cite{watanabe2018espnet} to generate the results for prior work.}\label{tab:tts}
%   \small
%   \begin{tabular}{lrrr|c}\toprule
%   &  \multicolumn{3}{c}{WER} & MOS~(95\%Confidence) \\
%   &LJSpeech  &VCTK &LibriTTS & VCTK \\\midrule
%   Tacotron2~\cite{shen2018natural} &4.4 &4.2 &7.3 &2.91$\pm$0.15 \\
%   FastSpeech2~\cite{ren2020fastspeech} &6.1 &3.8 &10.2 &3.03 $\pm$0.14\\
%   VITS~\cite{casanova2022yourtts} &6.4  &11.1 &8.3 &\textbf{3.56$\pm$0.12}\\
%   \ourstts (Ours) &\textbf{ 4.0} &\textbf{2.2} &\textbf{4.5} &3.34 $\pm$ 0.14 \\
%   \bottomrule
%   \end{tabular}
%   \vspace{-0.4cm}
%   \end{table}

  \begin{table}[!t]
\centering
\caption{WER (\%) (evaluated with WhisperX ASR ``base.en'') and MOS of different TTS models' generations using transcriptions from each evaluation set that correponds to data used for training.}\label{tab:tts}
\vspace{0.2cm}
\resizebox{0.7\textwidth}{!}{
\begin{tabular}{@{}lrrrc@{}}
\toprule
& \multicolumn{3}{c}{\textbf{WER↓ (\%)}} & \textbf{\begin{tabular}[c]{@{}c@{}}MOS↑ (95\% CI)\end{tabular}} \\
\cmidrule(lr){2-4} \cmidrule(l){5-5}
\textbf{Model} & \textbf{LJSpeech} &\textbf{LibriTTS} &
\textbf{VCTK} & 
\textbf{VCTK} \\
\midrule
GroundTruth & 2.6 & 3.8 & 3.4 & 4.18$\pm$0.10 \\ 
\midrule
Tacotron2, \cite{shen2018natural} & 4.4 &  7.3 & 4.2 & 2.91$\pm$0.15 \\
FastSpeech2, \cite{ren2020fastspeech} & 6.1 & 10.2 & 3.8 & 3.03$\pm$0.14 \\
VITS, \cite{casanova2022yourtts} & 6.4 & 8.3 & 11.1 & \textbf{3.56$\pm$0.12} \\
\ourstts (\dmel) & \textbf{4.0} & \textbf{4.5} & \textbf{2.2} & 3.34$\pm$0.14 \\
\bottomrule
\end{tabular}
}
% \vspace{-0.4cm}
\end{table}

\begin{table}[!t]
\centering
\caption{results for TTS models trained on LibriSpeech 960h and evaluated on LJSpeech test set.}\label{tab:tts-length}
\vspace{0.2cm}
\resizebox{0.6\textwidth}{!}{
\begin{tabular}{@{}lrrrr@{}}
\toprule
\textbf{Sequence Length} & \textbf{Tacotron2} & \textbf{FastSpeech2} & \textbf{VITS} & \textbf{\ourstts} \\
\midrule
Total WER↓ (\%) & 4.4 & 6.1 & 6.4 & \textbf{4.0} \\
10-20 words & 5.5 & \textbf{3.1} & 7.4 & 3.5 \\
20+ words & 3.3 & 9.1 & 5.3 & \textbf{3.0} \\
\bottomrule
\end{tabular}
}
% \vspace{-0.4cm}
\end{table}
  
  % Furthermore, we conductor human evaluation to measure the naturalness of the synthesized speech using the MOS score in Table~\ref{tab:mos-vctk}.

  % \begin{table}[!htp]\centering
  %   \caption{MOS score for TTS models on VCTK dataset}\label{tab:mos-vctk}
  %   \small
  %   \begin{tabular}{lr}\toprule
  %   &MOS \\\midrule
  %   GroundTruth &4.5 \\
  %   Tacotron2 &1.3 \\
  %   FastSpeech2 &1.3 \\
  %   VITS &4.1 \\
  %   Ours &3.9 \\
  %   \bottomrule
  %   \end{tabular}
  % \end{table}

  Furthermore, we observed that our model with \dmel tokenization can generate long audio sequences with high quality. Here, we evaluate the performance of our model on different lengths of text sequences using the LJSpeech test set.
  Table~\ref{tab:tts-length} shows the WER results for our model on text sequences with 10-20 words and more than 20 words. 
  We ignore text sequences with fewer than 10 words, as they are too short and not robust for WER evaluation.
  From Table~\ref{tab:tts-length}, we observe that our model achieves competitive performance across different text lengths, demonstrating its robustness and generalization ability in generating synthesized speech for varying text inputs lengths.
  Additionally, we find that the non-autoregressive (NAR) model FastSpeech2 achieves the lowest WER on shorter sequences but the highest WER on longer sequences. This suggests that NAR models may not be well-suited for generating long audio sequences.

% \begin{table}[!t]\centering
%   \caption{Text-to-speech results for models trained on Librispeech 960h data and evaluated on the transcriptions from LJSpeech test set. We show WER (\%) evaluated by WhisperX base.en ASR model on text grouped by sequence length.}\label{tab:tts-length}
%   \small
%   \begin{tabular}{lrrrrr}\toprule
%   WER &Tacotron2~\cite{shen2018natural} &FastSpeech2~\cite{ren2020fastspeech} &VITS(YourTTS)~\cite{casanova2022yourtts} &\ourstts (Ours) \\\midrule
%   total &4.4 &6.1 &6.4 & \textbf{4.0} \\
%   10-20 words &5.5 &\textbf{3.1} &7.4 &3.5 \\
%   more than 20 words &3.3 &9.1 &5.3 &\textbf{3.0 }\\
%   \bottomrule
%   \end{tabular}
%   % \vspace{-0.2cm}
%   \end{table}

\subsection{LM-Style Speech-to-Text}
% \begin{table}[!htp]\centering
%   \caption{ASR evaluation for speech tokenizer}\label{tab:main-asr}
%   \small
%   \begin{tabular}{lrrrrr}\toprule
%   &Dev-clean &Dev-other &Test-clean &Test-other \\\midrule
%   VOXTLM~\cite{maiti2024voxtlm} &- &- &6.5 &17.6 \\
%   SpeechTokenizer~\cite{zhang2023speechtokenizer} &8.08 &21.19 &8.6 &21.96 \\
%   HuBERT+KM &6.89 &17.44 &7.06 &17.59 \\
%   \dmel &\textbf{4.83} &\textbf{13.44} &\textbf{5.11} &\textbf{13.86} \\
%   \bottomrule
%   \end{tabular}
%   \end{table}

% \begin{table}[!t]\centering
%   \caption{Our Base ASR~(258M) models are trained on LibriSpeech 960h and evaluated for different speech tokenizers (greedy decoding is reported). We compare also with prior SOTA work on decoder-only model for ASR (VOXTLM). Results are shown across 2 runs with mean WER and standard deviation.}\label{tab:main-asr}
%   \small
%   \begin{tabular}{lllll}\toprule
%   &dev-clean &dev-other &test-clean &test-other \\\midrule
%   VOXTLM~\cite{maiti2024voxtlm} (HuBERT+KM~\cite{lakhotia2021generative}), 350M &- &- &6.5 &17.6 \\
%   VOXTLM~\cite{maiti2024voxtlm} (HuBERT+KM~\cite{lakhotia2021generative}), 1.3B &- &- &4.6 &12.1 \\
%   \midrule
%   \oursasr (SpeechTokenizer~\cite{zhang2023speechtokenizer}) & 6.5$\pm$0.3 & 16.9$\pm$0.7 & 6.9$\pm$0.4  & 17.5$\pm$0.5 \\
%   \oursasr (HuBERT+KM~\cite{lakhotia2021generative}) &5.3$\pm$0.1 & 13.7$\pm$0.2 & 5.8$\pm$0.1  & 13.8$\pm$0.1\\
%   \oursasr (\dmel) &\textbf{3.8}$\pm$0.1 & \textbf{10.3}$\pm$0.1 &\textbf{4.2}$\pm$0.2  &\textbf{10.4}$\pm$0.1 \\
%   \bottomrule
%   \end{tabular}
%   \vspace{-0.3cm}
%   \end{table}
\begin{table}[!t]
\centering
\caption{Speech recognition results for different tokenizers measured with WER (\%). All models are trained on LibriSpeech 960h.}\label{tab:main-asr}
\vspace{0.2cm}
\resizebox{0.8\textwidth}{!}{
\begin{tabular}{@{}lrrrrl@{}}
\toprule
\textbf{Model} & \textbf{dev-clean↓} & \textbf{dev-other↓} & \textbf{test-clean↓} & \textbf{test-other↓} & \textbf{Params} \\
\midrule
\oursasr (SpeechTokenizer) & 6.5$\pm$0.3 & 16.9$\pm$0.7 & 6.9$\pm$0.4 & 17.5$\pm$0.5 & 258M \\
\oursasr (HuBERT+KM) & 5.3$\pm$0.1 & 13.7$\pm$0.2 & 5.8$\pm$0.1 & 13.8$\pm$0.1 & 258M \\
\oursasr (\dmel) & \textbf{3.8}$\pm$0.1 & \textbf{10.3}$\pm$0.1 & \textbf{4.2}$\pm$0.2 & \textbf{10.4}$\pm$0.1 & 258M \\
\bottomrule
\end{tabular}
}
% \vspace{-0.4cm}
\end{table}

\begin{table}[!t]
\centering
\caption{Comparison of WER (\%) for best \oursasr trained with \dmel tokenization and prior work with LM-style ASR models and HuBERT+KM with subword modeling on top as tokenization.}\label{tab:main-asr-sota}
\vspace{0.2cm}
\resizebox{0.8\textwidth}{!}{
\begin{tabular}{@{}lcrrrrc@{}}
\toprule
\textbf{Model} & {\textbf{Data} (h)} & \textbf{dev-clean↓} & \textbf{dev-other↓} & \textbf{test-clean↓} & \textbf{test-other↓} & \textbf{Params} \\
\midrule
VOXTLM  & 280k & - & - & 6.5 & 17.6 & 350M \\
VOXTLM & 280k & - & - & 4.6 & 12.1 & 1.3B \\
Decoder-only ASR~\cite{chen2024loss} & 960 & 3.6 & \textbf{7.8} & 3.8 & \textbf{8.3} & 355M \\
\midrule
\oursasr (\dmel) & 960 & \textbf{3.1} & 8.4 & \textbf{3.4} & 8.6 & 355M \\
\bottomrule
\end{tabular}
}
% \vspace{-0.4cm}
\end{table}

Training an LM-style speech-to-text~(ASR) model can test if the speech tokens can preserve the semantic information in the speech signal and support the speech content-based task.
We use the same experiments configuration in Section\ref{sec:tts_config} to train ASR models.
Table~\ref{tab:main-asr} shows results of our model dubbed \oursasr, trained with different tokenizations including \dmel for the ASR task. Our LM-style model with \dmel speech tokenization achieves 4.2\% WER on the \textit{test-clean} and 10.4\% WER on the \textit{test-other} sets outperforming both HuBERT-KM and SpeechTokenizer.
We also observe that our model with HuBERT-KM~\citep{lakhotia2021generative} outperforms the SpeechTokenizer~\citep{zhang2023speechtokenizer} for ASR, which is reasonable as semantic tokens are more suitable for the ASR task.

In Table~\ref{tab:main-asr-sota}, we further compare \oursasr with \dmel speech tokenizer trained with GPT-2-meduim architecture~\citep{radford2019language} on LibriSpeech 960h with prior work: VOXTLM~\citep{maiti2024voxtlm} that uses larger model trained with more data and initialized from a pretrained LLM (OPT~\cite{zhang2022opt}), and HuBERT-KM with additional subword modeling on top as the speech tokenizer; \cite{chen2024loss} that also uses GPT-2 architecture trained on LibriSpeech 960h and HuBERT-KM with additional subword modeling on top as the speech tokenizer\footnote{We use official codebase to train this model w/o text pretraining as \cite{chen2024loss} report results only with text pretraining.}. 
\oursasr with \dmel outperforms VOXTLM; it also outperforms \cite{chen2024loss} on clean sets and a bit behind it on other sets.
% (we use exactly the same tokenization with 200 clusters)

The ASR results clearly demonstrate the benefit of using our \dmel speech tokenizer for the content-related tasks in speech, as it better preserves the semantic information in the speech signal. Further details and ablations can be found in Appendix~\ref{app:training-details} and~\ref{app:ablations}. 

\subsection{Ablations}
% Here we compare different \dmel variants by training and evaluating ASR and TTS models.

\begin{figure}[!t]
  \centering
  \begin{minipage}{0.48\textwidth}
    \centering
    \includegraphics[width=\textwidth]{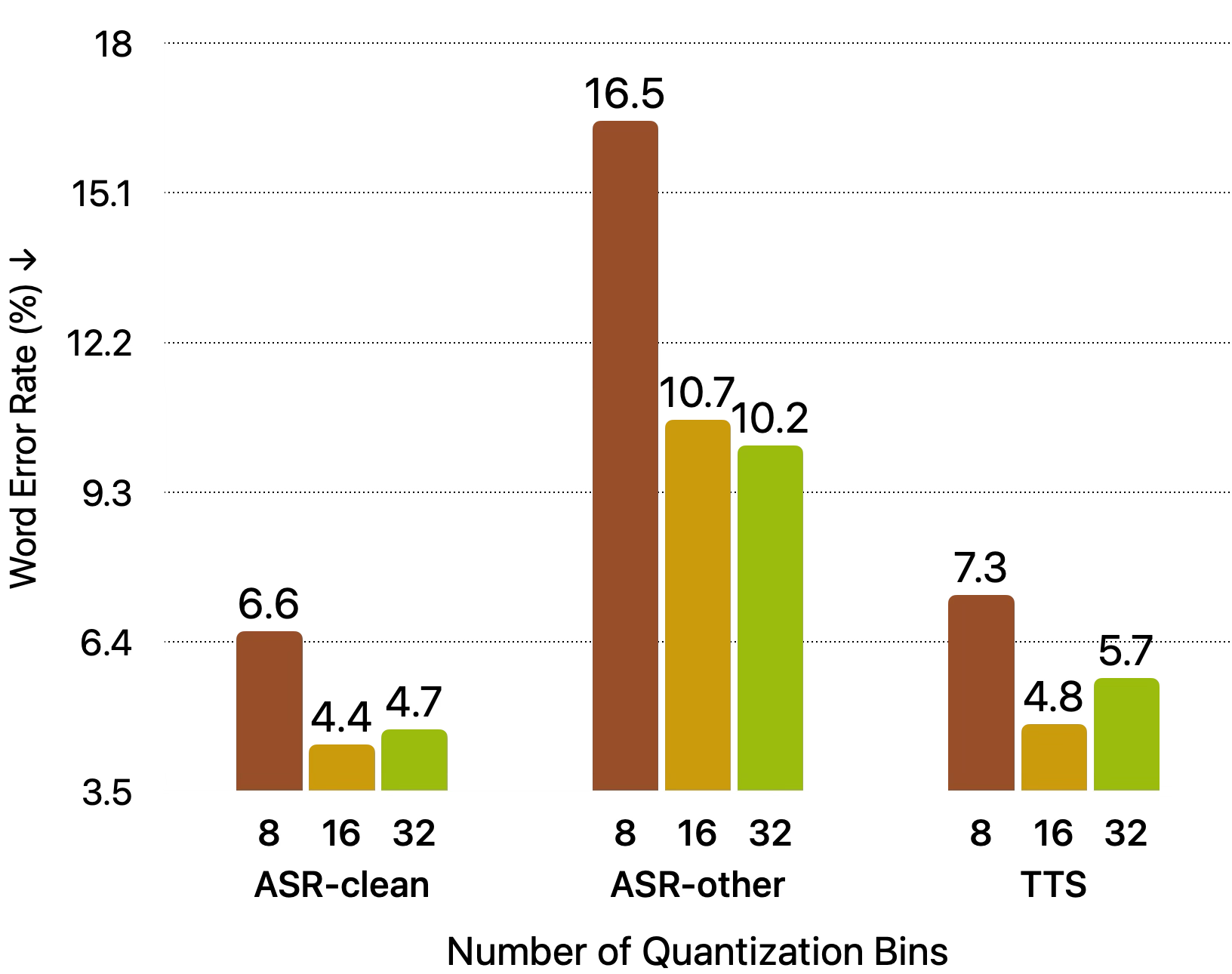}
    \caption{ASR and TTS results (WER, \%) with \dmel speech tokenizer and different number of bins~(codebook size) for discretization in \dmel. All models are trained on LibriSpeech 960h.}
    \label{fig:nbins-ablation}
  \end{minipage}
  \hfill
    \begin{minipage}{0.48\textwidth}
    \centering
    \includegraphics[width=\textwidth]{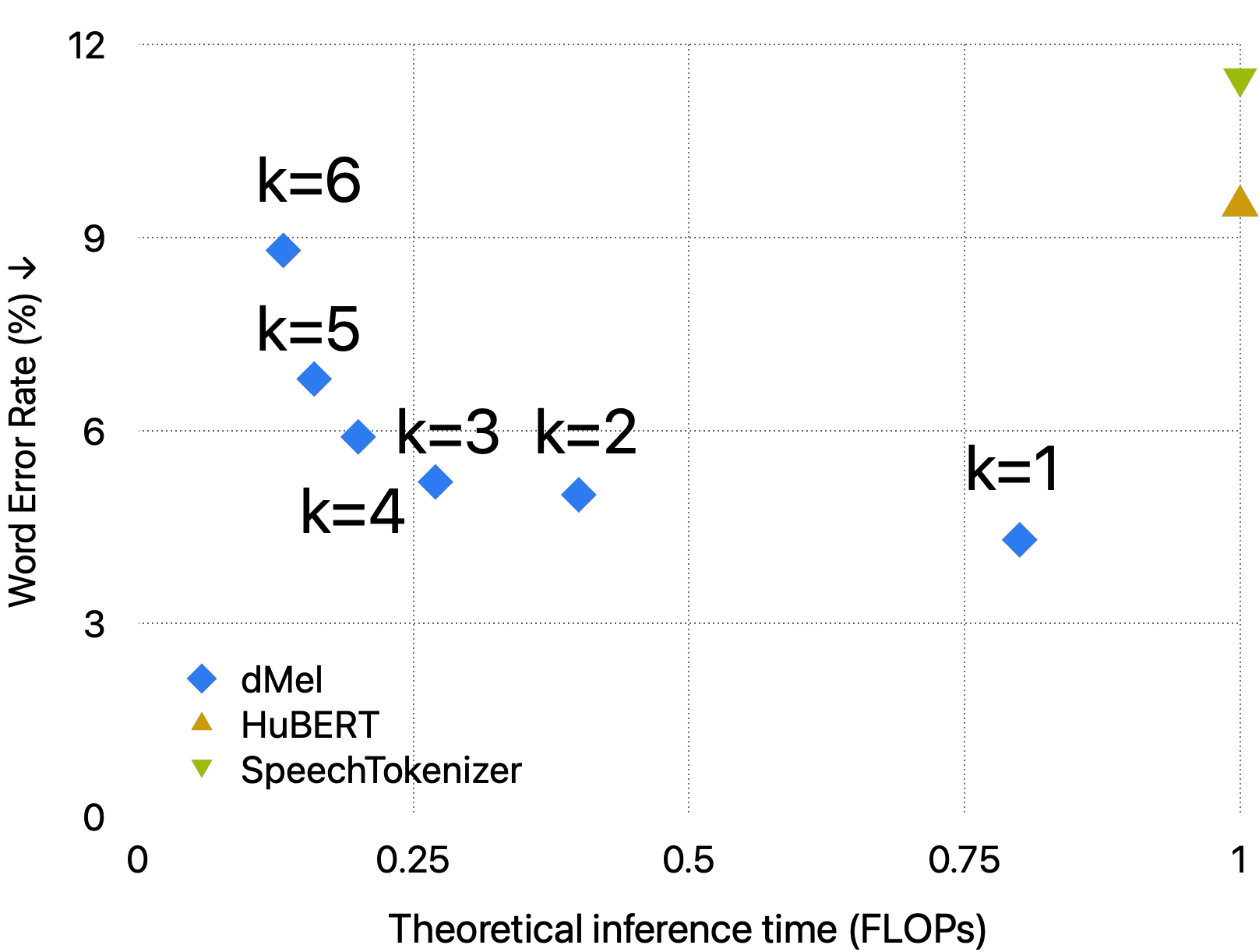}
    \caption{Ablation results on k-frame encoding and decoding.}
    \label{fig:richtts_multiframe_results}
  \end{minipage}

\end{figure}

We first investigate the impact of the codebook sizes, shown in Figure~\ref{fig:nbins-ablation}.
The 16-bin configuration used in the paper demonstrates the best overall performance across tasks. 
While the 32-bin setup slightly outperforms on the ASR \textit{test-other} set, it shows degraded
performance in TTS. 
This trade-off likely stems from the increased speech vocabulary size, which may pose challenges for accurate prediction. 
The results may get better with increased data and
model size. 
And 8-bin configuration looses too much information with discretization.

We then ablate the ASR results to understand why ASR LM-style model is behind the state-of-the-art on LibriSpeech. 
We take two existing transformer ASR baselines, Seq2Seq and CTC, that use 80 log mel-filterbanks and characters as targets.
We then modify these baselines by using \dmel instead (the discretization, embedding layer and linear layer) while keeping all other hyper-parameters the same (we adjust only the SpecAugment time masking max width accordingly to keep total masking in \textit{ms} the same).
Our results (Appendix Table~\ref{tab:asr-ablation}) suggest: i) \dmel brings only small degradation compared to Mel; ii) additional discrepancy is coming from different hop distance in featurization; iii) \textit{\textbf{the main and significant performance degradation is coming from switching to LM-style model.}}
The latter is in line with~\cite{maiti2024voxtlm} and~\cite{chen2024loss}, though was not discussed in detail by any prior work. We hypothesise this gap is due to observed overfitting of the LM-style models.

Finally, we conduct ablation on k-frame parallel encoding and decoding introduced in Section~\ref{sec:model}. 
The results are shown in Figure~\ref{fig:richtts_multiframe_results}.
As we can see from this figure, $k<=4$ yeild similar results to single-frame model, while improves both the training and inference efficiency significantly.
In contrast, for SpeechTokenizer, even $K=1$ is worse than \dmel with $k=6$.
This is because the residual nature of VQ tokens in SpeechTokenizer where each channel is a residual of the previous channel, which is not suitable for multi-channel parallel decoding.

\subsection{Unlocking Joint Speech-Text Modeling}

Our model design allows us to train a single model for both ASR and TTS tasks leading to a simpler setup.
We train a single model with the same architecture and tokenization as \ourstts, by constructing the training data with <text, speech> and <speech, text> pairs for ASR and TTS tasks, respectively.
By mixing these two types of data, we can train a single model for both tasks.

Table~\ref{tab:joint-asr} shows that the joint model is worse on both tasks, but ASR is affected more than TTS.
Comparing our results to VOXTLM, which initializes its model from pretrained LLM (OPT) and finetunes it with multiple tasks and datasets, we speculate that our joint model needs text-only training to learn a good LM for better ASR performance. Our model structure trivially allows for this text-only training, but we leave those experiments for future work (for further discussion see Appendix~\ref{app:joint}).

\section{Related Work}
% \vspace{-0.1cm}
\paragraph{Speech Tokenization} Recent advancements in speech tokenization have primarily focused on two approaches: semantic tokens and acoustic tokens. This section examines these methods, their combinations, and their limitations, highlighting the need for more efficient and generalizable solutions.
Semantic tokens, extracted from self-supervised pretrained speech models, have shown promise in capturing high-level content information. Methods like wav2vec \citep{baevski2020wav2vec} and HuBERT \citep{hsu2021hubert} employ $k$-means clustering on speech representations to generate these tokens. 
While effective in capturing semantic content, these approaches often struggle with preserving fine-grained acoustic details crucial for high-quality speech synthesis.
In contrast, acoustic tokens, derived from pretrained audio compression models, excel at preserving low-level acoustic information. 
Techniques such as SoundStream \citep{zeghidour2021soundstream} and EnCodec \citep{defossez2022high} utilize residual vector quantization (RVQ) with reconstruction objectives. These methods achieve high-quality audio compression but may not capture higher-level semantic structures effectively.

Recognizing the complementary nature of semantic and acoustic tokens, recent works have attempted to combine these approaches. 
AudioLM \citep{borsos2023audiolm} introduced a three-stage model: semantic modeling, coarse acoustic modeling, and fine acoustic modeling. While comprehensive, this approach introduces complexity and computational overhead. 
AudioPalm \citep{rubenstein2023audiopalm} further demonstrated the critical importance of large-scale training data and model parameters for effective multi-stage modeling, highlighting potential generalization issues in low-resource scenarios.
An alternative hybrid approach, proposed by \citet{zhang2023speechtokenizer}, attempts to distill semantic information into acoustic tokens during RVQ model training. 
However, this method still requires additional pretraining and does not fully achieve a single-stage model architecture.

Despite these advancements, several challenges persist in the field of speech tokenization: i) balancing semantic and acoustic information in a unified representation; ii) reducing model complexity and computational requirements; iii) improving generalization to low-resource / out-of-domain data e.g. with mixed speech from multiple speakers, or multiple languages, or changing characteristics of recording equipment/sampling rate etc.; iv) developing truly single-stage tokenizers.
Our proposed method, \dmel, addresses these challenges by offering a training-free speech tokenization approach. 
By directly discretizing log mel-filterbanks into bins, it inherently preserves both semantic and acoustic information in a unified representation, while significantly reducing computational complexity.
Concurrently, \cite{langman2024spectral} proposed a different mel-filterbanks based speech tokens: spectral codecs, where disjointed mel-bands are encoded separately and then quantized using an FSQ~\citep{mentzer2023finite}. 
Although spectral codecs and \dmel are both discretizing mel-filterbanks, \dmel is encoder-free and tested with autoregressive generation tasks, while spectral codecs need encoders and is tested with non-autogressive generation task.
Also, our scalar quantization method shares some similarities with FSQ, but FSQ is in a learned latent code space and necessitates an additional bound operation to limit the range of the latent codes.

\begin{table}[!tp]\centering
\caption{Results of ASR and TTS jointly trained model on LibriSpeech 960h.} \label{tab:joint-asr}
\vspace{0.2cm}
\small
\resizebox{0.9\textwidth}{!}{
\begin{tabular}{lrrrrr}\toprule
&\multicolumn{2}{c}{\textbf{ASR, WER↓ (\%)}} &\multicolumn{2}{c}{\textbf{TTS}} \\\cmidrule{2-5}
&\textbf{test-clean} & \textbf{test-other} & \textbf{WER↓ (\%)} &\textbf{CER↓ (\%)} \\\midrule
VOXTLM+OPT, 350M &3.5 &8.7 &- &3.5 \\
\oursasr-\ourstts, single models, 258M &4.2 & 10.4 &4.3 &1.8 \\
% \oursasr-\ourstts, joint model, 258M & 7.6 &20.0 &4.4 &1.9 \\
\oursasr-\ourstts, joint model, 258M & 7.5 &15.3 &4.4 & 1.9 \\
\bottomrule
\end{tabular}
}
% \vspace{-0.4cm}
\end{table}
% \vspace{-0.4cm}
\vspace{-0.2cm}
\paragraph{Speech-Text Modeling} Modeling speech and text jointly is a challenging task, as speech signals are continuous and while text is discrete. 
Existing works have explored various approaches to address this challenge, including usage of separate encoders for different modalities~\citep{ao2021speecht5,bapna2021slam}.
\citet{bai20223} proposed an encoder-only model A3T for speech-text modeling, by introducing alignment embedding to encourage cross-modal transfer between text and speech.
Although A3T achieved good performance on speech synthesis and editing tasks, it cannot generate text and cannot generalize to longform generation because of its encoder-only architecture and mask-reconstruction training strategy.
VioLA~\citep{wang2023viola} also targets a unified speech-text model which can generate speech and text with a single model, but it is specifically designed for the Encodec~\citep{defossez2022high} style feature, and compelled to model speech tokens in a multi-stage hierarchical manner.
\citet{maiti2024voxtlm} proposed a LM-style model VOXTLM, to model speech and text jointly. 
However, VOXTLM is only models the HuBERT semantic tokens, and relies on an external generation model to transform semantic tokens into waveform, but the speaker and acoustic information are lost.
In comparison, the model architecture in this paper is a simple, single stage LM-style transformer model, and can handle both the speech generation and text generation tasks.

% \hl{TBD~\cite{kim2024clamtts}}
\section{Conclusion}
In this work, we proposed \dmel, \textit{a novel train-free speech tokenization} method that discretizes log mel-filterbank energies directly into bins. \textit{By operating on the authentic log mel-filterbank representation, \dmel inherently generalizes to out-of-domain data (e.g. speech with noise or other languages), preserves both semantic and acoustic information in a unified tokenized representation, and is streamable.} Our key contribution is the evaluation of \dmel within a unified LM-style transformer architecture for speech recognition (ASR) and speech synthesis (TTS) tasks. 
Our \dmel-based ASR model, \oursasr, achieved the lowest word error rate among tokenization methods, robustly preserving semantic content. For TTS, \dmel's generation yielded the lowest WER, accurately reconstructing speech waveforms. Our \dmel-based TTS model, \ourstts, achieved competitive naturalness, lowest error rates, and long audio generation capabilities.

\dmel's simplicity circumvents separate tokenizers or multi-stage modeling, reducing computational overhead and dependence on pretrained models. By unifying semantic and acoustic modeling, \dmel enables efficient speech-text modeling frameworks.
While initial joint TTS-ASR training showed promise, further work is needed. Our primary contribution demonstrates \dmel's effectiveness for high-performing separate TTS and ASR models within a unified LM-style architecture.

% \clearpage

\balance
\bibliographystyle{plainnat}
\bibliography{references}

\begin{thebibliography}{52}
\providecommand{\natexlab}[1]{#1}
\providecommand{\url}[1]{\texttt{#1}}
\expandafter\ifx\csname urlstyle\endcsname\relax
  \providecommand{\doi}[1]{doi: #1}\else
  \providecommand{\doi}{doi: \begingroup \urlstyle{rm}\Url}\fi

\bibitem[Ao et~al.(2021)Ao, Wang, Zhou, Wang, Ren, Wu, Liu, Ko, Li, Zhang, et~al.]{ao2021speecht5}
Junyi Ao, Rui Wang, Long Zhou, Chengyi Wang, Shuo Ren, Yu~Wu, Shujie Liu, Tom Ko, Qing Li, Yu~Zhang, et~al.
\newblock Speecht5: Unified-modal encoder-decoder pre-training for spoken language processing.
\newblock \emph{arXiv preprint arXiv:2110.07205}, 2021.

\bibitem[Baevski et~al.(2020)Baevski, Zhou, Mohamed, and Auli]{baevski2020wav2vec}
Alexei Baevski, Yuhao Zhou, Abdelrahman Mohamed, and Michael Auli.
\newblock wav2vec 2.0: A framework for self-supervised learning of speech representations.
\newblock \emph{Advances in neural information processing systems}, 33:\penalty0 12449--12460, 2020.

\bibitem[Bai et~al.(2022)Bai, Zheng, Chen, Ma, Li, and Huang]{bai20223}
He~Bai, Renjie Zheng, Junkun Chen, Mingbo Ma, Xintong Li, and Liang Huang.
\newblock A$^3${T}: Alignment-aware acoustic and text pretraining for speech synthesis and editing.
\newblock In \emph{International Conference on Machine Learning}, pages 1399--1411. PMLR, 2022.

\bibitem[Bain et~al.(2023)Bain, Huh, Han, and Zisserman]{bain2022whisperx}
Max Bain, Jaesung Huh, Tengda Han, and Andrew Zisserman.
\newblock Whisperx: Time-accurate speech transcription of long-form audio.
\newblock \emph{INTERSPEECH 2023}, 2023.

\bibitem[Bapna et~al.(2021)Bapna, Chung, Wu, Gulati, Jia, Clark, Johnson, Riesa, Conneau, and Zhang]{bapna2021slam}
Ankur Bapna, Yu-an Chung, Nan Wu, Anmol Gulati, Ye~Jia, Jonathan~H Clark, Melvin Johnson, Jason Riesa, Alexis Conneau, and Yu~Zhang.
\newblock Slam: A unified encoder for speech and language modeling via speech-text joint pre-training.
\newblock \emph{arXiv preprint arXiv:2110.10329}, 2021.

\bibitem[Bengio et~al.(2015)Bengio, Vinyals, Jaitly, and Shazeer]{scheduled_sampling}
Samy Bengio, Oriol Vinyals, Navdeep Jaitly, and Noam Shazeer.
\newblock Scheduled sampling for sequence prediction with recurrent neural networks.
\newblock \emph{Advances in neural information processing systems}, 28, 2015.

\bibitem[Bogdanov et~al.(2019)Bogdanov, Won, Tovstogan, Porter, and Serra]{bogdanov2019mtg}
Dmitry Bogdanov, Minz Won, Philip Tovstogan, Alastair Porter, and Xavier Serra.
\newblock The mtg-jamendo dataset for automatic music tagging.
\newblock In \emph{ML4MD Machine Learning for Music Discovery Workshop, ICML}, 2019.

\bibitem[Borsos et~al.(2023)Borsos, Marinier, Vincent, Kharitonov, Pietquin, Sharifi, Roblek, Teboul, Grangier, Tagliasacchi, et~al.]{borsos2023audiolm}
Zal{\'a}n Borsos, Rapha{\"e}l Marinier, Damien Vincent, Eugene Kharitonov, Olivier Pietquin, Matt Sharifi, Dominik Roblek, Olivier Teboul, David Grangier, Marco Tagliasacchi, et~al.
\newblock Audiolm: a language modeling approach to audio generation.
\newblock \emph{IEEE/ACM Transactions on Audio, Speech, and Language Processing}, 2023.

\bibitem[Brown et~al.(2020)Brown, Mann, Ryder, Subbiah, Kaplan, Dhariwal, Neelakantan, Shyam, Sastry, Askell, et~al.]{brown2020language}
Tom Brown, Benjamin Mann, Nick Ryder, Melanie Subbiah, Jared~D Kaplan, Prafulla Dhariwal, Arvind Neelakantan, Pranav Shyam, Girish Sastry, Amanda Askell, et~al.
\newblock Language models are few-shot learners.
\newblock \emph{Advances in neural information processing systems}, 33:\penalty0 1877--1901, 2020.

\bibitem[Casanova et~al.(2022)Casanova, Weber, Shulby, Junior, G{\"o}lge, and Ponti]{casanova2022yourtts}
Edresson Casanova, Julian Weber, Christopher~D Shulby, Arnaldo~Candido Junior, Eren G{\"o}lge, and Moacir~A Ponti.
\newblock Yourtts: Towards zero-shot multi-speaker tts and zero-shot voice conversion for everyone.
\newblock In \emph{International Conference on Machine Learning}, pages 2709--2720. PMLR, 2022.

\bibitem[Chen et~al.(2024)Chen, Wang, Zhang, Zheng, Zhang, Deng, Ma, Yu, Liu, and Zhang]{chen2024loss}
Qian Chen, Wen Wang, Qinglin Zhang, Siqi Zheng, Shiliang Zhang, Chong Deng, Yukun Ma, Hai Yu, Jiaqing Liu, and Chong Zhang.
\newblock Loss masking is not needed in decoder-only transformer for discrete-token-based asr.
\newblock In \emph{ICASSP 2024-2024 IEEE International Conference on Acoustics, Speech and Signal Processing (ICASSP)}, pages 11056--11060. IEEE, 2024.

\bibitem[D{\'e}fossez et~al.(2022)D{\'e}fossez, Copet, Synnaeve, and Adi]{defossez2022high}
Alexandre D{\'e}fossez, Jade Copet, Gabriel Synnaeve, and Yossi Adi.
\newblock High fidelity neural audio compression.
\newblock \emph{arXiv preprint arXiv:2210.13438}, 2022.

\bibitem[D{\'e}fossez et~al.(2024)D{\'e}fossez, Mazar{\'e}, Orsini, Royer, P{\'e}rez, J{\'e}gou, Grave, and Zeghidour]{defossez2024moshi}
Alexandre D{\'e}fossez, Laurent Mazar{\'e}, Manu Orsini, Am{\'e}lie Royer, Patrick P{\'e}rez, Herv{\'e} J{\'e}gou, Edouard Grave, and Neil Zeghidour.
\newblock Moshi: a speech-text foundation model for real-time dialogue.
\newblock \emph{arXiv preprint arXiv:2410.00037}, 2024.

\bibitem[Dong et~al.(2018)Dong, Xu, and Xu]{dong2018speech}
Linhao Dong, Shuang Xu, and Bo~Xu.
\newblock Speech-transformer: a no-recurrence sequence-to-sequence model for speech recognition.
\newblock In \emph{2018 IEEE international conference on acoustics, speech and signal processing (ICASSP)}, pages 5884--5888. IEEE, 2018.

\bibitem[Graves et~al.(2006)Graves, Fern{\'a}ndez, Gomez, and Schmidhuber]{graves2006connectionist}
Alex Graves, Santiago Fern{\'a}ndez, Faustino Gomez, and J{\"u}rgen Schmidhuber.
\newblock Connectionist temporal classification: labelling unsegmented sequence data with recurrent neural networks.
\newblock In \emph{Proceedings of the 23rd International Conference on Machine Learning}, pages 369--376, 2006.

\bibitem[Gulati et~al.(2020)Gulati, Qin, Chiu, Parmar, Zhang, Yu, Han, Wang, Zhang, Wu, et~al.]{gulati2020conformer}
Anmol Gulati, James Qin, Chung-Cheng Chiu, Niki Parmar, Yu~Zhang, Jiahui Yu, Wei Han, Shibo Wang, Zhengdong Zhang, Yonghui Wu, et~al.
\newblock Conformer: Convolution-augmented transformer for speech recognition.
\newblock \emph{arXiv preprint arXiv:2005.08100}, 2020.

\bibitem[Hines et~al.(2012)Hines, Skoglund, Kokaram, and Harte]{hines2012visqol}
Andrew Hines, Jan Skoglund, Anil Kokaram, and Naomi Harte.
\newblock Visqol: The virtual speech quality objective listener.
\newblock In \emph{IWAENC 2012; international workshop on acoustic signal enhancement}, pages 1--4. VDE, 2012.

\bibitem[Hsu et~al.(2021)Hsu, Bolte, Tsai, Lakhotia, Salakhutdinov, and Mohamed]{hsu2021hubert}
Wei-Ning Hsu, Benjamin Bolte, Yao-Hung~Hubert Tsai, Kushal Lakhotia, Ruslan Salakhutdinov, and Abdelrahman Mohamed.
\newblock Hubert: Self-supervised speech representation learning by masked prediction of hidden units.
\newblock \emph{IEEE/ACM Transactions on Audio, Speech, and Language Processing}, 29:\penalty0 3451--3460, 2021.

\bibitem[Ito and Johnson(2017)]{ljspeech17}
Keith Ito and Linda Johnson.
\newblock The lj speech dataset.
\newblock \url{https://keithito.com/LJ-Speech-Dataset/}, 2017.

\bibitem[Kim et~al.(2021)Kim, Kong, and Son]{kim2021conditional}
Jaehyeon Kim, Jungil Kong, and Juhee Son.
\newblock Conditional variational autoencoder with adversarial learning for end-to-end text-to-speech.
\newblock In \emph{International Conference on Machine Learning}, pages 5530--5540. PMLR, 2021.

\bibitem[Kim et~al.(2024)Kim, Lee, Chung, and Cho]{kim2024clamtts}
Jaehyeon Kim, Keon Lee, Seungjun Chung, and Jaewoong Cho.
\newblock {CL}am-{TTS}: Improving neural codec language model for zero-shot text-to-speech.
\newblock In \emph{The Twelfth International Conference on Learning Representations}, 2024.
\newblock URL \url{https://openreview.net/forum?id=ofzeypWosV}.

\bibitem[Kim et~al.(2022)Kim, Gholami, Shaw, Lee, Mangalam, Malik, Mahoney, and Keutzer]{kim2022squeezeformer}
Sehoon Kim, Amir Gholami, Albert Shaw, Nicholas Lee, Karttikeya Mangalam, Jitendra Malik, Michael~W Mahoney, and Kurt Keutzer.
\newblock Squeezeformer: An efficient transformer for automatic speech recognition.
\newblock \emph{Advances in Neural Information Processing Systems}, 35:\penalty0 9361--9373, 2022.

\bibitem[Kong et~al.(2020)Kong, Kim, and Bae]{kong2020hifi}
Jungil Kong, Jaehyeon Kim, and Jaekyoung Bae.
\newblock Hifi-gan: Generative adversarial networks for efficient and high fidelity speech synthesis.
\newblock \emph{Advances in neural information processing systems}, 33:\penalty0 17022--17033, 2020.

\bibitem[Lakhotia et~al.(2021)Lakhotia, Kharitonov, Hsu, Adi, Polyak, Bolte, Nguyen, Copet, Baevski, Mohamed, et~al.]{lakhotia2021generative}
Kushal Lakhotia, Eugene Kharitonov, Wei-Ning Hsu, Yossi Adi, Adam Polyak, Benjamin Bolte, Tu-Anh Nguyen, Jade Copet, Alexei Baevski, Abdelrahman Mohamed, et~al.
\newblock On generative spoken language modeling from raw audio.
\newblock \emph{Transactions of the Association for Computational Linguistics}, 9:\penalty0 1336--1354, 2021.

\bibitem[Langman et~al.(2024)Langman, Juki{\'c}, Dhawan, Koluguri, and Ginsburg]{langman2024spectral}
Ryan Langman, Ante Juki{\'c}, Kunal Dhawan, Nithin~Rao Koluguri, and Boris Ginsburg.
\newblock Spectral codecs: Spectrogram-based audio codecs for high quality speech synthesis.
\newblock \emph{arXiv preprint arXiv:2406.05298}, 2024.

\bibitem[Le et~al.(2023)Le, Vyas, Shi, Karrer, Sari, Moritz, Williamson, Manohar, Adi, Mahadeokar, and Hsu]{le2023voicebox}
Matthew Le, Apoorv Vyas, Bowen Shi, Brian Karrer, Leda Sari, Rashel Moritz, Mary Williamson, Vimal Manohar, Yossi Adi, Jay Mahadeokar, and Wei-Ning Hsu.
\newblock Voicebox: Text-guided multilingual universal speech generation at scale, 2023.

\bibitem[Lee et~al.(2023)Lee, Ping, Ginsburg, Catanzaro, and Yoon]{leebigvgan}
Sang-gil Lee, Wei Ping, Boris Ginsburg, Bryan Catanzaro, and Sungroh Yoon.
\newblock Bigvgan: A universal neural vocoder with large-scale training.
\newblock In \emph{The Eleventh International Conference on Learning Representations}, 2023.

\bibitem[Maiti et~al.(2024)Maiti, Peng, Choi, Jung, Chang, and Watanabe]{maiti2024voxtlm}
Soumi Maiti, Yifan Peng, Shukjae Choi, Jee-weon Jung, Xuankai Chang, and Shinji Watanabe.
\newblock Voxtlm: Unified decoder-only models for consolidating speech recognition, synthesis and speech, text continuation tasks.
\newblock In \emph{ICASSP 2024-2024 IEEE International Conference on Acoustics, Speech and Signal Processing (ICASSP)}, pages 13326--13330. IEEE, 2024.

\bibitem[Mentzer et~al.(2023)Mentzer, Minnen, Agustsson, and Tschannen]{mentzer2023finite}
Fabian Mentzer, David Minnen, Eirikur Agustsson, and Michael Tschannen.
\newblock Finite scalar quantization: Vq-vae made simple.
\newblock \emph{arXiv preprint arXiv:2309.15505}, 2023.

\bibitem[Mousavi et~al.(2024)Mousavi, Della~Libera, Duret, Ploujnikov, Subakan, and Ravanelli]{mousavi2024dasb}
Pooneh Mousavi, Luca Della~Libera, Jarod Duret, Artem Ploujnikov, Cem Subakan, and Mirco Ravanelli.
\newblock Dasb--discrete audio and speech benchmark.
\newblock \emph{arXiv preprint arXiv:2406.14294}, 2024.

\bibitem[Panayotov et~al.(2015)Panayotov, Chen, Povey, and Khudanpur]{panayotov2015librispeech}
Vassil Panayotov, Guoguo Chen, Daniel Povey, and Sanjeev Khudanpur.
\newblock Librispeech: an asr corpus based on public domain audio books.
\newblock In \emph{2015 IEEE international conference on acoustics, speech and signal processing (ICASSP)}, pages 5206--5210. IEEE, 2015.

\bibitem[Park et~al.(2019)Park, Chan, Zhang, Chiu, Zoph, Cubuk, and Le]{DBLP:conf/interspeech/ParkCZCZCL19}
Daniel~S. Park, William Chan, Yu~Zhang, Chung{-}Cheng Chiu, Barret Zoph, Ekin~D. Cubuk, and Quoc~V. Le.
\newblock Specaugment: {A} simple data augmentation method for automatic speech recognition.
\newblock In Gernot Kubin and Zdravko Kacic, editors, \emph{Interspeech 2019, 20th Annual Conference of the International Speech Communication Association, Graz, Austria, 15-19 September 2019}, pages 2613--2617. {ISCA}, 2019.
\newblock \doi{10.21437/INTERSPEECH.2019-2680}.
\newblock URL \url{https://doi.org/10.21437/Interspeech.2019-2680}.

\bibitem[Radford et~al.(2019)Radford, Wu, Child, Luan, Amodei, Sutskever, et~al.]{radford2019language}
Alec Radford, Jeffrey Wu, Rewon Child, David Luan, Dario Amodei, Ilya Sutskever, et~al.
\newblock Language models are unsupervised multitask learners.
\newblock \emph{OpenAI blog}, 1\penalty0 (8):\penalty0 9, 2019.

\bibitem[Radford et~al.(2023)Radford, Kim, Xu, Brockman, McLeavey, and Sutskever]{radford2023robust}
Alec Radford, Jong~Wook Kim, Tao Xu, Greg Brockman, Christine McLeavey, and Ilya Sutskever.
\newblock Robust speech recognition via large-scale weak supervision.
\newblock In \emph{International Conference on Machine Learning}, pages 28492--28518. PMLR, 2023.

\bibitem[Raffel et~al.(2020)Raffel, Shazeer, Roberts, Lee, Narang, Matena, Zhou, Li, and Liu]{raffel2020exploring}
Colin Raffel, Noam Shazeer, Adam Roberts, Katherine Lee, Sharan Narang, Michael Matena, Yanqi Zhou, Wei Li, and Peter~J Liu.
\newblock Exploring the limits of transfer learning with a unified text-to-text transformer.
\newblock \emph{Journal of machine learning research}, 21\penalty0 (140):\penalty0 1--67, 2020.

\bibitem[Ren et~al.(2020)Ren, Hu, Tan, Qin, Zhao, Zhao, and Liu]{ren2020fastspeech}
Yi~Ren, Chenxu Hu, Xu~Tan, Tao Qin, Sheng Zhao, Zhou Zhao, and Tie-Yan Liu.
\newblock Fastspeech 2: Fast and high-quality end-to-end text to speech.
\newblock \emph{arXiv preprint arXiv:2006.04558}, 2020.

\bibitem[Rouditchenko et~al.(2024)Rouditchenko, Collobert, and Likhomanenko]{anonymous2024avcpl}
Andrew Rouditchenko, Ronan Collobert, and Tatiana Likhomanenko.
\newblock {AV}-{CPL}: Continuous pseudo-labeling for audio-visual speech recognition.
\newblock In \emph{ECCV 2024 Workshop - AVGenL: Audio-Visual Generation and Learning}, 2024.

\bibitem[Rubenstein et~al.(2023)Rubenstein, Asawaroengchai, Nguyen, Bapna, Borsos, Quitry, Chen, Badawy, Han, Kharitonov, et~al.]{rubenstein2023audiopalm}
Paul~K Rubenstein, Chulayuth Asawaroengchai, Duc~Dung Nguyen, Ankur Bapna, Zal{\'a}n Borsos, F{\'e}lix de~Chaumont Quitry, Peter Chen, Dalia~El Badawy, Wei Han, Eugene Kharitonov, et~al.
\newblock Audiopalm: A large language model that can speak and listen.
\newblock \emph{arXiv preprint arXiv:2306.12925}, 2023.

\bibitem[Shen et~al.(2018)Shen, Pang, Weiss, Schuster, Jaitly, Yang, Chen, Zhang, Wang, Skerrv-Ryan, et~al.]{shen2018natural}
Jonathan Shen, Ruoming Pang, Ron~J Weiss, Mike Schuster, Navdeep Jaitly, Zongheng Yang, Zhifeng Chen, Yu~Zhang, Yuxuan Wang, Rj~Skerrv-Ryan, et~al.
\newblock Natural tts synthesis by conditioning wavenet on mel spectrogram predictions.
\newblock In \emph{2018 IEEE international conference on acoustics, speech and signal processing (ICASSP)}, pages 4779--4783. IEEE, 2018.

\bibitem[Shi et~al.(2022)Shi, Hsu, Lakhotia, and Mohamed]{shi2022learning}
Bowen Shi, Wei-Ning Hsu, Kushal Lakhotia, and Abdelrahman Mohamed.
\newblock Learning audio-visual speech representation by masked multimodal cluster prediction.
\newblock In \emph{International Conference on Learning Representations}, 2022.
\newblock URL \url{https://openreview.net/forum?id=Z1Qlm11uOM}.

\bibitem[Su et~al.(2024)Su, Ahmed, Lu, Pan, Bo, and Liu]{su2024roformer}
Jianlin Su, Murtadha Ahmed, Yu~Lu, Shengfeng Pan, Wen Bo, and Yunfeng Liu.
\newblock Roformer: Enhanced transformer with rotary position embedding.
\newblock \emph{Neurocomputing}, 568:\penalty0 127063, 2024.

\bibitem[Toyin(2024)]{toyin2024unified}
Hawa Toyin.
\newblock A unified model for text-to-speech and speech-to-text.
\newblock 2024.

\bibitem[Variani et~al.(2014)Variani, Lei, McDermott, Moreno, and Gonzalez-Dominguez]{variani2014deep}
Ehsan Variani, Xin Lei, Erik McDermott, Ignacio~Lopez Moreno, and Javier Gonzalez-Dominguez.
\newblock Deep neural networks for small footprint text-dependent speaker verification.
\newblock In \emph{2014 IEEE international conference on acoustics, speech and signal processing (ICASSP)}, pages 4052--4056. IEEE, 2014.

\bibitem[Wang et~al.(2023{\natexlab{a}})Wang, Chen, Wu, Zhang, Zhou, Liu, Chen, Liu, Wang, Li, et~al.]{wang2023neural}
Chengyi Wang, Sanyuan Chen, Yu~Wu, Ziqiang Zhang, Long Zhou, Shujie Liu, Zhuo Chen, Yanqing Liu, Huaming Wang, Jinyu Li, et~al.
\newblock Neural codec language models are zero-shot text to speech synthesizers.
\newblock \emph{arXiv preprint arXiv:2301.02111}, 2023{\natexlab{a}}.

\bibitem[Wang et~al.(2023{\natexlab{b}})Wang, Zhou, Zhang, Wu, Liu, Gaur, Chen, Li, and Wei]{wang2023viola}
Tianrui Wang, Long Zhou, Ziqiang Zhang, Yu~Wu, Shujie Liu, Yashesh Gaur, Zhuo Chen, Jinyu Li, and Furu Wei.
\newblock Viola: Unified codec language models for speech recognition, synthesis, and translation.
\newblock \emph{arXiv preprint arXiv:2305.16107}, 2023{\natexlab{b}}.

\bibitem[Yamagishi et~al.(2019)Yamagishi, Veaux, and MacDonald]{vctk2019}
Junichi Yamagishi, Christophe Veaux, and Kirsten MacDonald.
\newblock Cstr vctk corpus: English multi-speaker corpus for cstr voice cloning toolkit (version 0.92).
\newblock University of Edinburgh. The Centre for Speech Technology Research (CSTR). \url{https://datashare.ed.ac.uk/handle/10283/2950}, 2019.

\bibitem[Yamamoto et~al.(2020)Yamamoto, Song, and Kim]{yamamoto2020parallel}
Ryuichi Yamamoto, Eunwoo Song, and Jae-Min Kim.
\newblock Parallel wavegan: A fast waveform generation model based on generative adversarial networks with multi-resolution spectrogram.
\newblock In \emph{ICASSP 2020-2020 IEEE International Conference on Acoustics, Speech and Signal Processing (ICASSP)}, pages 6199--6203. IEEE, 2020.

\bibitem[Zeghidour et~al.(2021)Zeghidour, Luebs, Omran, Skoglund, and Tagliasacchi]{zeghidour2021soundstream}
Neil Zeghidour, Alejandro Luebs, Ahmed Omran, Jan Skoglund, and Marco Tagliasacchi.
\newblock Soundstream: An end-to-end neural audio codec.
\newblock \emph{IEEE/ACM Transactions on Audio, Speech, and Language Processing}, 30:\penalty0 495--507, 2021.

\bibitem[Zen et~al.(2019)Zen, Dang, Clark, Zhang, Weiss, Jia, Chen, and Wu]{zen2019libritts}
Heiga Zen, Viet Dang, Rob Clark, Yu~Zhang, Ron~J Weiss, Ye~Jia, Zhifeng Chen, and Yonghui Wu.
\newblock Libritts: A corpus derived from librispeech for text-to-speech.
\newblock \emph{arXiv preprint arXiv:1904.02882}, 2019.

\bibitem[Zhang et~al.(2023{\natexlab{a}})Zhang, Li, Zhang, Zhan, Wang, Zhou, and Qiu]{zhang2023speechgpt}
Dong Zhang, Shimin Li, Xin Zhang, Jun Zhan, Pengyu Wang, Yaqian Zhou, and Xipeng Qiu.
\newblock Speechgpt: Empowering large language models with intrinsic cross-modal conversational abilities.
\newblock \emph{arXiv preprint arXiv:2305.11000}, 2023{\natexlab{a}}.

\bibitem[Zhang et~al.(2022)Zhang, Roller, Goyal, Artetxe, Chen, Chen, Dewan, Diab, Li, Lin, et~al.]{zhang2022opt}
Susan Zhang, Stephen Roller, Naman Goyal, Mikel Artetxe, Moya Chen, Shuohui Chen, Christopher Dewan, Mona Diab, Xian Li, Xi~Victoria Lin, et~al.
\newblock Opt: Open pre-trained transformer language models.
\newblock \emph{arXiv preprint arXiv:2205.01068}, 2022.

\bibitem[Zhang et~al.(2023{\natexlab{b}})Zhang, Zhang, Li, Zhou, and Qiu]{zhang2023speechtokenizer}
Xin Zhang, Dong Zhang, Shimin Li, Yaqian Zhou, and Xipeng Qiu.
\newblock Speechtokenizer: Unified speech tokenizer for speech large language models.
\newblock \emph{arXiv preprint arXiv:2308.16692}, 2023{\natexlab{b}}.

\end{thebibliography}

%%%%%%%%%%%%%%%%%%%%%%%%%%%%%%%%%%%%%%%%%%%%%%%%%%%%%%%%%%%%
\newpage
\appendix

\section{Ethics Statement}
\label{ethic}
% TBD write about it as we use generative models

% From VoiceBox example
% As with other powerful new AI innovations, we recognize this technology brings the potential for misuse and unintended harm. In our paper, we detail how we built a highly effective classifier that can distinguish between authentic speech and audio generated with Voicebox to mitigate these possible future risks. There are many exciting use cases for generative speech models, but because of the risks of misuse, we are not making the Voicebox model or code publicly available at this time. While we believe it is important to be open with the AI community and to share our research to advance the state of the art in AI, it’s also necessary to strike the right balance between openness with responsibility.
The development and deployment of speech technologies carry important ethical considerations. While our proposed \dmel method aims to advance the state-of-the-art in speech-text modeling, it is crucial to highlight potential ethical risks and raise the awareness so that new methods may be developed to mitigate these risks.

Our first main concern is the potential dual-use of speech synthesis technologies for nefarious purposes such as impersonation, misleading audio-visual content generation, or voice spoofing attacks.
Proactive measures, including watermarking techniques and robust speaker verification methods, should be explored to counter such risks.
The former attempts to build markers into the generated speech that make it easy to detect, while the latter focusses on distinguishing synthetic from real data.
Prior work~\citep{le2023voicebox} has shown that neural networks can be trained to distinguish speech synthesized from their model from real speech, probably because of artifacts from the use of mel spectral vocoders.
While we did not train a network to do so in our work yet (we will create one before code release), the vocoders we use are similar to their work -- going from mel spectrogram to raw waveforms. 
Our model also does not use prosody, phoneme duration and other predictions that more sophisticated TTS systems use to allow the model to perform very well on imitating speaker styles in zero-shot settings.
However our model can probably mimic the styles of training speakers very well.
It is our hope that releasing our methods will facilitate more research on fake speech verification and watermarking techniques -- even if current classifiers are able to perform this detection, the quality of the generative models is improving.
It is also our hope that future works will attempt to perform more credit assignment -- by providing metrics that show which real data samples a synthetic speech example copies its style and substance from.

Another concern is the perpetuation of societal biases encoded within training data. Speech datasets may exhibit biases along dimensions such as gender, race, age, or socioeconomic status, which could be propagated or amplified by trained models. Rigorous debiasing techniques and careful curation of representative training data are essential to mitigate these risks. On the mitigating side of this equation, we also hope that with better, more controllable TTS systems, ASR systems can improve because more data can be generated for underrepresented segments of the distribution from the TTS models.

Furthermore, the development and deployment of speech technologies should prioritize accessibility and inclusivity. Models should be evaluated for performance across diverse demographics, accents, and language varieties to ensure equitable access and quality of service.

Finally, it is important to foster transparency and accountability in the research and development process. Clear documentation of model capabilities, limitations, and potential failure modes should be provided to enable informed decision-making and responsible usage.

Addressing these ethical considerations requires a multistakeholder approach involving researchers, developers, policymakers, and end-users. By prioritizing ethical principles such as fairness, privacy, and accountability, we can work towards realizing the benefits of speech technologies while mitigating potential risks and adverse societal impacts.

% \section{Limitations}\label{app:limitations}

% Because TTS work is tremendously fragmented and clear protocols are not often available for training and evaluation, we reimplemented other tokenizers within our code base using publicly available, official implementations where available. 
% We trained our models on those tokenizations. While we made the best effort to tune the tokenization methods and the models, there is always a possibility we missed some details. However, our results seems to tell a consistent story when viewed from multiple angles, and when viewed on multiple datasets. For the joint model, we did not do extensive multi-task training using text only data and we intend to do that in the future work.
% We also did not train on larger model sizes (>1B parameters), larger datasets (>1k hours), or using pretrained models.

\section{Limitations}\label{app:limitations}

Because TTS work is tremendously fragmented and clear protocols are not often available for training and evaluation, we reimplemented other tokenizers within our code base using publicly available, official implementations where available: e.g. we used Hubert-KM and speech tokenizer features extraction from the public codebases and pluged them into our LM-style model training. 
% We trained our models on those tokenizations. 
While we made the best effort to tune the tokenization methods and the models, there is always a possibility we missed some details. 
However, our results seem to tell a consistent story when viewed from multiple angles, and when viewed on multiple datasets. 
% For the joint model, we did not do extensive multi-task training using text only data and we intend to do that in the future work.
We also did not train on larger model sizes (>1B parameters), larger datasets (>1k hours), or using pretrained models.

The real challenge for modern multimodal LLMs is complex semantic understanding tasks.
While our current experiments focus on text-to-speech and speech-to-text tasks, these encompass critical aspects of speech processing. \dmel's effective performance within a decoder-only architecture for both tasks suggests potential for broader applications. 
We recognize the importance of more sophisticated speech understanding tasks and view our work as a foundation for future research leaving other tasks out of scope of the paper. 
Scaling up pretraining and exploring complex semantic understanding tasks could further validate our approach's versatility across a wider range of multimodal language processing challenges.

We acknowledge that our current scope targets only speech on purpose, as indicated in our title. 
While \dmel may potentially support non-speech tasks, \textbf{\textit{our current exploration and verification focus solely on speech, not general audio.}} 
Regarding the ``speaker variations'' -- mel-spectrogram is used for speaker recognition widely, thus it preserves necessary speaker information on which we thus rely in \dmel too.

% \section{Acknowledgements}\label{app:ack}

% We thank Dan Busbridge, Ronan Collobert and Jason Ramapuram for their helpful feedback and critical discussions at all stages of the research; Rick Chang, David Grangier, Barry Theobald for their helpful feedback on the initial paper draft. Names are in alphabetical order by last name within group.

\section{Data, Code, Reproducibility}\label{app:repro}
We made the best effort to use publicly available data and official implementations of prior works where it is possible. All data we used are under permissive license for research. We provided as much as detail as is possible without code such as details on our model training and hyperparameters throughout the paper and in the Appendix. We plan to open-source our code upon paper acceptance.

We do not plan to open-source any pre-trained models for sake of privacy, safety and misuse.

\section{Subjective Evaluation for TTS}\label{subjective_evals}
We use crowd-sourcing to collect subjective ratings to compare the naturalness of the reconstructed speech from the different tokenizers. We evaluate the quality of the same (randomly sampled) 50 utterances for each model by collecting around seven ratings per sample. Overall, we collect  3500 ratings from 65 raters. The raters were English-speaking and were paid at least the minimum wage. 

\begin{figure}[!h]
  \centering
  % \vspace{-0.4cm}
  \includegraphics[width=0.8\textwidth]{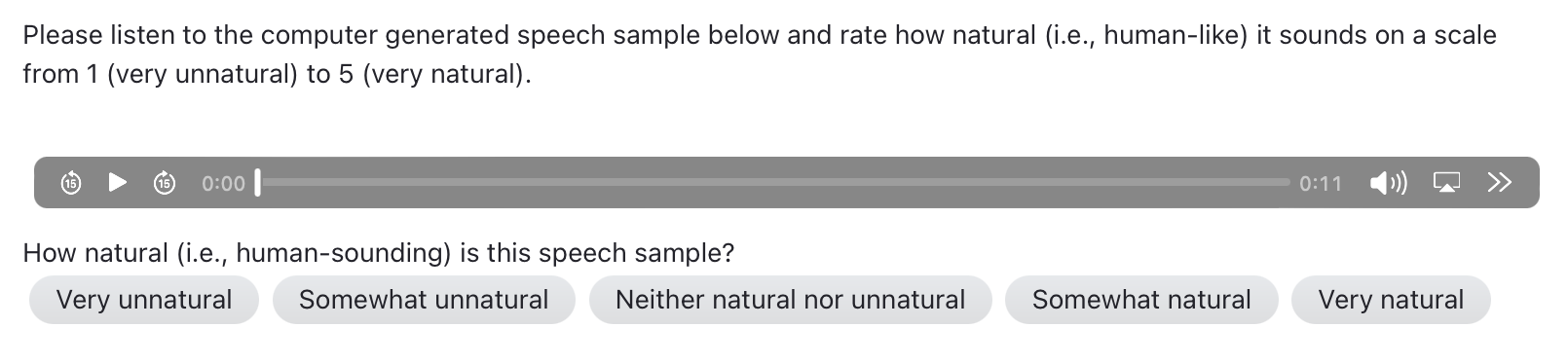}
  % \vspace{-0.5cm}
  \caption{A screenshot of the assessment task, as the crowd-sourced rater sees it.}
  \label{fig:crowd_screenshot}
\end{figure}

We present the raters with a generated speech sample and instruct them to rate how natural it sounds on a five-point Likert scale, where 1 corresponds to very unnatural and 5 corresponds to very natural. Figure~\ref{fig:crowd_screenshot} shows a screenshot of our subjective test as seen by the rater.

We noticed human annotators have bias over audio volume so we do volume normalization on top of all reconstructed or generated audio before giving them to human annotators.

We report Mean Opinion Score (MOS) results throughout the paper with confidence intervals calculated using bootstrap resampling with 1000 iterations, providing a reliable estimate of the variability MOS results.

\section{Training Details}\label{app:training-details}

\subsection{Baselines}
For reproducibility, we provide the HuggingFace model cards used in our experiments in Table~\ref{tab:tts}:
\begin{itemize}[leftmargin=0.6cm]
    \item Tacotron2 \citep{shen2018natural}, \url{https://huggingface.co/espnet/espnet/kan-bayashi_vctk_tts_train_xvector_tacotron2_raw_phn_tacotron_g2p_en_no_space_train.loss.ave}
    \item FastSpeech2 \citep{ren2020fastspeech}, \url{https://huggingface.co/espnet/kan-bayashi_vctk_gst_fastspeech2}
    \item VITS~\citep{casanova2022yourtts}, \url{https://huggingface.co/espnet/kan-bayashi_vctk_multi_spk_vits}
    \item ParallelWaveGAN~\citep{yamamoto2020parallel}, \url{https://github.com/kan-bayashi/ParallelWaveGAN/blob/master/egs/libritts/voc1/conf/parallel_wavegan.v1.yaml}
    \item HifiGAN~\citep{kong2020hifi}, \url{https://github.com/kan-bayashi/ParallelWaveGAN/blob/master/egs/libritts/voc1/conf/hifigan.v1.yaml}
    \item BigVGAN~\citep{leebigvgan}, \url{https://huggingface.co/nvidia/bigvgan_24khz_100band}
\end{itemize}

\subsection{\oursasr and \ourstts}

For our LM-style model we stack together speaker embedding, speech tokens and text tokens. Both speech and text tokens have prepended begin of sentence token (<bos>) and appended end of sentence token (<eos>).

We train all models using the Adam optimizer with a learning rate of 1e-3, learning rate warmup of 4k steps for ASR and 5k for TTS, cosine learning rate schedule and gradient clipping of 1.0 for TTS and 0.1 for ASR and joint models.
We use dynamic batching to optimize the data packing with total batch size of 1.4h/1.4h/0.7h for ASR training and 1h/2h/2h for TTS training for Small/Base/Large models.
We train TTS models for 100k steps and ASR models 80k steps with mixed precision training and BF16 on A100 and H100 GPUs with 80GB.
Both ASR models and TTS models are trained with 8GPUs for less than a day and for 2-4 days for ASR and TTS respectively.

% table of model sizes
\begin{table}[h]\centering
    \caption{LM-style transformer model configurations for ASR, TTS and joint models training.}\label{tab:model-size}
    \small
    \vspace{0.2cm}
    \begin{tabular}{lrrr}\toprule
    &\textbf{Small} &\textbf{Base} &\textbf{Large} \\\midrule
    \# of layers &18 &36 &48 \\
    \# of attention heads &2 &4 &8 \\
    \# of hidden units ($D$) &512 &768 &1536 \\
    \midrule
    \# of parameters &59M &258M &1.3B \\
    \bottomrule
    \end{tabular}
    \vspace{-0.2cm}
    \end{table}

\subsection{LM-Style Speech-to-Text}

For ASR training as an augmentation we apply SpecAugment~\citep{DBLP:conf/interspeech/ParkCZCZCL19} with 2 frequency masks with max width 30 and 10 time masks with max width 50 and ratio 0.1. 
With ablations we found that SpecAugment masking with average value instead of zero is slightly better.
Without applying SpecAugment performance of ASR is 7.3\% WER on \textit{dev-clean} and 20.3\% WER on \textit{dev-other}, which is further can be improved with usage of frequency masking only to 6.4\% WER on \textit{dev-clean} and 16.6\% WER on \textit{dev-other}. Usage of both frequency masking and time masking results in the best performance of Table~\ref{tab:main-asr}.

% fixed in the init
% During experiments with ASR decoder-only models and \dmel tokenization we observed some model training instabilities resulting in the spikes in the gradient norm and thus sometimes spikes in training loss.
% To stabilize training i) we reduce the gradient clipping from 1.0 used for TTS training to 0.1; ii) we add queries and keys normalizations via LayerNorm~\cite{dehghani2023scaling} on the head dimension before computing attention matrix.

We found that span masking is key part of model training to enforce self-attention to attend to speech part as well as to reduce exposure bias. The masking strategy is similar to the one used for TTS training: for every training step with probability $p$ the sample in the minibatch is masked with the mean span of 3 tokens with masking ration of 0.5. We found that the mean span of 1 token or 5 tokens gives the same results; while the mask probability $p$ is the most important hyper-parameter. The optimimal value for ASR is found to be 0.8, which is used in all final models.

As we found one best model configuration for the Base model with \dmel we then change only i) model size ii) speech tokenization iii) training data (here we increase model dropout to 0.3 for training on \textit{train-clean-360} and to 0.5 for training on \textit{train-clean-100} as otherwise models drastically overfit); the rest of hyper-parameters stay the same.

\section{Ablations}\label{app:ablations}

\subsection{LM-Style Text-to-Speech}
Scaling results for \ourstts are shown in Table~\ref{tab:abl-tts}.

\begin{table}[!t]\centering
  \caption{Text-to-speech results for different model sizes with \dmel. All models are trained on LibriSpeech 960h dataset. Evaluation is done via speech generation on the full \textit{test-clean} transcriptions and speakers, and then evaluated WER with WhisperX base.en.}\label{tab:abl-tts}
  \small
  \vspace{0.2cm}
  \begin{tabular}{lrr}\toprule
  &\textbf{WER↓ (\%)} \\\midrule
  \ourstts (\dmel), Small &8.1 \\
  \ourstts (\dmel), Base & 4.3 \\
  \ourstts (\dmel), Large & 5.4 \\
  \bottomrule
\end{tabular}
% \vspace{-0.4cm}
\end{table}

\subsection{LM-Style Speech-to-Text}

ASR ablations for different model sizes, data sizes, and tokenizers are shown in Table~\ref{tab:extended-asr}.

We noticed the results in \cite{chen2024loss} seems to be the SOTA for LM-style ASR model to the best of our knowledge.
However, as many ablations are missed in~\cite{chen2024loss}, we took their open-sourced code and run ablations ourselves to have proper comparison with it. The final results, including ablation with \dmel are shown in Table~\ref{tab:chen-ablation}:
\begin{itemize}[leftmargin=0.6cm]
    \item We successfully reproduced~\cite{chen2024loss} results (row 1 and 2).
    \item Without pretraining (rows 3, 4, 5):
    
        \dmel outperforms HuBERT-KM on both clean and other datasets;
        \dmel surpasses BPE on top of HuBERT-KM on clean data, while BPE on HuBERT-KM performs better on other.
    \item Without pretraining and without speed perturbation (rows 6, 7, 8):
    
        BPE on HuBERT-KM performance decreases significantly after diabling speed perturbation (compare rows 3 and 6), raising questions about its generalizability to other domains, given that BPE tokens are trained on speed-perturbed LibriSpeech data.
        
        Our \dmel (row 8) achieves substantially better results than both HuBERT-KM and BPE on HuBERT-KM (rows 7 and 6), demonstrating robust performance even without speed augmentation.
\end{itemize}

Note that in \dmel, we use SpecAugment (masking across time and channels) and \cite{chen2024loss} also use SpecAugment. According to their code, the time masking is 30\%, while channel masking is impossible as there is only 1 channel).

We believe these results demonstrate the effectiveness, simplicity in use, and robustness of our \dmel tokenization method, particularly in scenarios where extensive pretraining or domain-specific augmentations may not be feasible.

Note that~\cite{chen2024loss} did not show applicability of BPE on HuBERT-KM or HuBERT-KM to TTS task, while in VOXTLM (also uses BPE on HuBERT-KM) it is shown that this tokenization is not suited for TTS (the performance is poor). \dmel in contrary is shown to perform well on TTS task too in addition to ASR. 

\begin{table}[!tp]\centering
\caption{WER (\%) comparison for CTC, Seq2Seq, and LM-style ASR models ($\sim$260M) trained on LibriSpeech 960h with \dmel and Mel features. We compute 80 log mel-filterbanks with 25ms (50ms) window and 10ms (25ms) hop distance, denoted as `10ms' (`25ms').}\label{tab:asr-ablation}
\small
\vspace{0.2cm}
\begin{tabular}{lrccc}\toprule
\textbf{Model} & \textbf{Features} & \textbf{dev-clean↓} & \textbf{dev-other↓} \\
\toprule
\cite{gulati2020conformer} (RNN-T -- Conformer) & Mel-10ms & 1.9 & 4.1 \\
\cite{kim2022squeezeformer} (CTC -- Squeezeformer) & Mel-10ms & 2.3 & 5.8 \\
\midrule
\multirow{4}{*}{Seq2Seq~\citep{dong2018speech}} & Mel-10ms & 2.4 & 5.4 \\
& \dmel-10ms & 2.5 & 5.9 \\
& Mel-25ms & 2.8 & 6.5  \\
& \dmel-25ms & 2.7 & 6.2  \\
\midrule
\multirow{4}{*}{CTC~\citep{graves2006connectionist}} & Mel-10ms & 2.1 & 5.4 \\
& \dmel-10ms & 2.1 & 5.6 \\ 
& Mel-25ms & 2.1 & 5.4 \\
& \dmel-25ms & 2.3 & 6.1 \\
\midrule
LM-style & \dmel-25ms & 3.4 & 9.5 \\
\bottomrule
\end{tabular}
\vspace{-0.4cm}
\end{table}

\begin{table}[!tp]\centering
  \caption{Our ASR models trained on different subsets (\textit{train-clean-100} LS-100, \textit{train-clean-360} LS-360, full LibriSpeech LS-960) of LibriSpeech, with different model sizes and different speech tokenizations (greedy decoding is reported). Results are shown across 2 runs with mean WER and standard deviation.}\label{tab:extended-asr}
  \vspace{0.2cm}
  \resizebox{1\textwidth}{!}{
  \begin{tabular}{lccrrrr}\toprule
  \textbf{Tokenization} & \textbf{Model Size} & \textbf{Data} & \textbf{dev-clean↓} &\textbf{dev-other↓} &\textbf{test-clean↓} &\textbf{test-other↓} \\\midrule
  \dmel & \multirow{2}{*}{Base} & LS-100 & 18.1$\pm$1.0 & 39.4$\pm$1.2 &19.0 $\pm$1.0  & 41.3$\pm$1.1 \\
  \dmel & & LS-360 & 6.4$\pm$0.4 & 20.1$\pm$1.1 &6.9 $\pm$0.6  & 20.5$\pm$0.9 \\
  \midrule
  \midrule
  SpeechTokenizer~\citep{zhang2023speechtokenizer} & \multirow{3}{*}{Small} & \multirow{3}{*}{LS-960} & 6.2$\pm$0.2 & 16.8$\pm$0.3 & 6.5$\pm$0.2  & 17.4$\pm$0.3 \\
  HuBERT+KM~\citep{lakhotia2021generative} & & &5.8$\pm$0.2 & 14.6$\pm$0.1 & 6.0$\pm$0.1  & 14.9$\pm$0.1\\
  \dmel & & & 6.0$\pm$0.4 & 15.2$\pm$0.8 &6.1$\pm$0.4  & 15.7$\pm$0.7 \\
  \midrule
  \midrule
  SpeechTokenizer~\citep{zhang2023speechtokenizer} & \multirow{3}{*}{Base} & \multirow{3}{*}{LS-960} & 6.5$\pm$0.3 & 16.9$\pm$0.7 & 6.9$\pm$0.4  & 17.5$\pm$0.5 \\
  HuBERT+KM~\citep{lakhotia2021generative} & & &5.3$\pm$0.1 & 13.7$\pm$0.2 & 5.8$\pm$0.1  & 13.8$\pm$0.1\\
  \dmel & & &\textbf{3.8}$\pm$0.1 & \textbf{10.3}$\pm$0.1 &\textbf{4.2}$\pm$0.2  &\textbf{10.4}$\pm$0.1 \\
  \bottomrule
  \end{tabular}
  }
  \end{table}

\begin{table}[!tp]\centering
  \caption{Ablations on the LM-style ASR model with GPT-2 architecture using the setup from~\cite{chen2024loss}: we ablate pretraining with text, speed perturbation and speech tokenization methods. All models are trained on LibriSpeech 960h. We report WER (\%) on LibriSpeech validation and test sets.}\label{tab:chen-ablation}
  \vspace{0.2cm}
  \resizebox{1\textwidth}{!}{
  \begin{tabular}{lccrrrr}\toprule
  \textbf{Tokenization} & \textbf{Text Pretraining} & \textbf{Speed Perturbation} & \textbf{dev-clean↓} &\textbf{dev-other↓} &\textbf{test-clean↓} &\textbf{test-other↓} \\\midrule
  BPE on HuBERT+KM, \cite{chen2024loss}  	& \checkmark	& \checkmark & 2.9 &	6.2 	& 3.0 & 	6.6 \\
 BPE on HuBERT+KM, reproduction  	& \checkmark &	\checkmark &	2.9 &	6.3 &	3.2 &	6.7 \\ 
 \midrule
 BPE on HuBERT+KM &	\xmark & \checkmark	& 3.6 	& 7.8 & 	3.8  &	8.3 \\
 HuBERT+KM 	& 	\xmark & \checkmark	& 5.1 	& 8.9 &	5.5 	& 9.3 \\
 \dmel & 	\xmark & 	\checkmark	& 3.1 	& 8.4 	& 3.4 	& 8.6 \\
 \midrule
 BPE on HuBERT+KM  &	\xmark & \xmark & 	5.4 &	9.7 &	5.3 &	10.0 \\
 HuBERT+KM 	& \xmark	& \xmark & 	6.0 &	10.5 	&6.4 	&10.9 \\
\dmel & \xmark & \xmark &	3.7  &	9.7 	& 3.9 &	9.5 \\
  \bottomrule
  \end{tabular}
  }
  % \vspace{-0.4cm}
  \end{table}
  
\subsection{Joint Speech-Text Modeling Discussion}\label{app:joint}

We found it to be challenging to train joint model for ASR and TTS, similar to observations as in~\cite{maiti2024voxtlm} and e.g. \cite{shi2022learning,anonymous2024avcpl} for joint audio-visual speech recognition. Also, there is a very recent research work~\citep{toyin2024unified}, that also shows training TTS and ASR jointly is challenging, and needs carefully designed model architecture and training loss fusion technique. 

One of the reasons is the different pace of learning. Careful consideration of training strategies can mitigate some of the challenges in joint modeling of TTS and ASR tasks, highlighting the complexities inherent in combining these distinct but related tasks within a single model.

Another reason we suspect is the mismatch between train and test time, which is more pronounced for the joint modeling: if we compare individual validation losses per task in joint model to their one-task training counterparts we see they match each other (so training is fine), however the generation (test time which mismatches how the train loss is defined) for both tasks is broken: longer sequences has hallucination and high repetition issues. This could be due to different length of sequences between text and audio and thus learnt attention pattern could be different which creates longer sequences generation issue for the joint model.

Last but not the least, the two tasks have opposite modalities in the input and output, making it rather difficult to model. Most previously researched multi-task work have the same modality in the output. The combination of ASR and TTS is a rather recent phenomenon, such as Viola and VOXTLM.
%%%%%%%%%%%%%%%%%%%%%%%%%%%%%%%%%%%%%%%%%%%%%%%%%%%%%%%%%%%%

\clearpage

\end{document}